\documentclass{article}

\PassOptionsToPackage{numbers, sort&compress, comma, square}{natbib}
\usepackage[final]{neurips_2022}

\usepackage[utf8]{inputenc} % allow utf-8 input
\usepackage[T1]{fontenc}    % use 8-bit T1 fonts
\usepackage{hyperref}       % hyperlinks
\usepackage{url}            % simple URL typesetting
\usepackage{booktabs}       % professional-quality tables
\usepackage{amsfonts}       % blackboard math symbols
\usepackage{nicefrac}       % compact symbols for 1/2, etc.
\usepackage{microtype}      % microtypography
\usepackage{xcolor}         % colors

\usepackage{xspace}
\usepackage{tabularx}
\usepackage{wrapfig,lipsum}
\usepackage{graphicx} 
\usepackage{multirow}
\usepackage{CJK}
\usepackage{colortbl}
\usepackage{caption}
\usepackage{subcaption}
\usepackage{enumitem}
\usepackage{latexsym}
\usepackage{amsmath}
\usepackage{arydshln}

\usepackage{lettrine}
\usepackage{algorithm} 
\usepackage[noend]{algpseudocode}
\usepackage[normalem]{ulem}
\newcommand{\method}{\textsc{CoNT}~}

\title{\textsc{CoNT}: Contrastive Neural Text Generation}

\author{Chenxin An$^{1,2}$\thanks{This work was done during Chenxin An’s internship at Shanghai AI Laboratory}, Jiangtao Feng$^{2}$, Kai Lv$^{1}$, Lingpeng Kong$^{2,3}$, Xipeng Qiu$^{1}$, Xuanjing Huang$^{1,4}$ \\
  $^1$Fudan University, $^2$Shark-NLP Shanghai AI Laboratory \\
  $^3$The University of Hong Kong \\
  $^4$Shanghai Collaborative Innovation Center of Intelligent Visual Computing \\
  \texttt{\{cxan20, klv21, xpqiu, xjhuang\}@fudan.edu.cn}\\
  \texttt{fengjiangtao@pjlab.org.cn}, \texttt{lpk@cs.hku.hk}\\
 }

\begin{document}
\maketitle
\begin{abstract}
Recently, contrastive learning attracts increasing interests in neural text generation as a new solution to alleviate the \textit{exposure bias} problem. 
It introduces a sequence-level training signal which is crucial to generation tasks that always rely on auto-regressive decoding. 
However, previous methods using contrastive learning in neural text generation usually lead to inferior performance.
In this paper, we analyse the underlying reasons and propose a new \textbf{Co}ntrastive \textbf{N}eural \textbf{T}ext generation framework, \textsc{CoNT}. 
\textsc{CoNT} addresses bottlenecks that prevent contrastive learning from being widely adopted in generation tasks from three aspects -- the construction of contrastive examples, the choice of the contrastive loss, and the strategy in decoding.
We validate \textsc{CoNT} on five generation tasks with ten benchmarks, including machine translation, summarization, code comment generation, data-to-text generation and commonsense generation. 
Experimental results show that \textsc{CoNT} clearly outperforms the conventional training framework on all the ten benchmarks with a convincing margin. 
Especially, \textsc{CoNT} surpasses previous the most competitive contrastive learning method for text generation, by 1.50 BLEU on machine translation and 1.77 ROUGE-1 on summarization, respectively.
It achieves new state-of-the-art on summarization, code comment generation (without external data) and data-to-text generation. \footnote{The code is available at \url{https://github.com/Shark-NLP/CoNT} }

\end{abstract}

% body part
\section{Introduction}
Contrastive learning has achieved great success in representation learning \cite{van2017neural, van2018representation, chen2020simple}. It also attracts enormous interests in neural text generation recently. By creating positive and negative examples in response to unseen (or erroneous) inputs \cite{lee2020contrastive}, contrastive learning offers a new solution to alleviate the \emph{exposure bias} problem \cite{bengio2015scheduled, paulus2017deep} -- an autoregressive model trained only using the ground truths exhibits inferior generalization performance. Apart from that, contrastive learning also introduces a sequence-level loss in addition to the conventional token-level language model loss with maximum likelihood estimation (MLE). This is crucial to most conditional text generation tasks (e.g., machine translation and summarization) which are evaluated on sequence-level metrics (e.g., BLEU~\cite{papineni2002bleu}). 

However, it is non-trivial to get contrastive learning working on neural text generation. If we simply use from-batch positive-negative samples following simCLR~\cite{chen2020simple}, and adopt the InfoNCE loss \cite{van2018representation, he2020momentum} which ignores the difference between negative samples (\S\ref{sec:naive-cl}; Naive CL), the improvement over non-contrastive baselines on generation tasks is rather marginal. Previous work attempts to build better contrastive samples by disturbing the ground truth \cite{ssmba, gao2021simcse, lee2020contrastive} in the discrete space or the continuous embedding space, but when it comes to text generation tasks, their performance gains are still far from satisfactory.

\begin{figure}[t!]
\centering
    \includegraphics[width=0.9\linewidth]{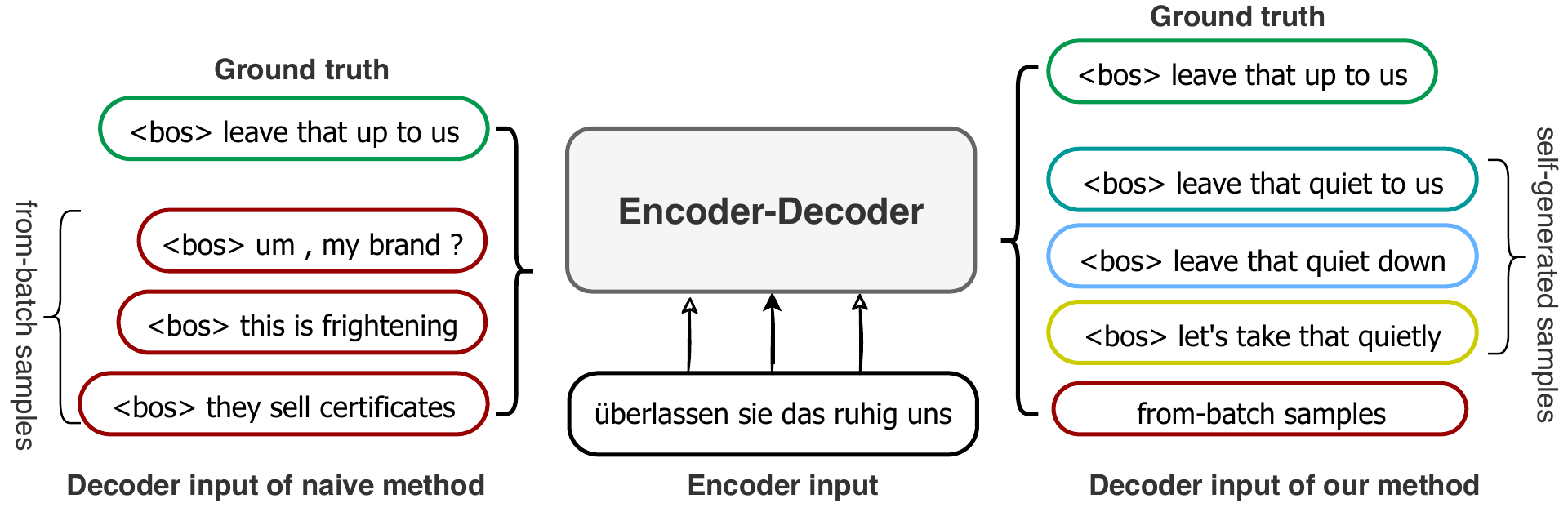}
\vspace{-0.5em}
\caption{A case study from IWSLT'14 De-En translation task. The naive setting uses from-batch samples following SimCLR~\cite{chen2020simple}. Compared with the naive method, \textsc{CoNT} both incorporates self-generated samples and from-batch samples. The border color indicates the acutual distance between the ground truth and the contrastive example.}
\label{fig:model}
\end{figure}

In this work, we propose a new contrastive neural text generation framework, \textsc{CoNT}.  \textsc{CoNT} does three different things from previous frameworks that make suboptimal use of contrastive learning. First, \textsc{CoNT} samples contrastive examples from its own predictions (e.g., through  beam search algorithm). This training procedure exposes the model to its mistakes in the inference stage and effectively alleviate the exposure bias problem. We show a comparison between negative samples in \textsc{CoNT} and in Naive CL in Figure~\ref{fig:model}. 
Second, we use a N-pairs contrastive loss which gives a fine-grained treatment to the contrastive examples based on their sequence-level scores (e.g., BLEU). It allows the model to fully leverage the supervision from the ground truth example (and its own generated examples) to learn a better sequence-level distance function between the source and the target representation. Third, we directly incorporate the learned sequence similarity score from the distance function into the inference stage. This helps the model to find a better global configuration, than merely follows the language model likelihood 
objective in decoding.

We validate \textsc{CoNT} on various important conditional language generation tasks (\S\ref{sec:mt}), including machine translation, summarization, code comment generation, data-to-text generation, and commonsense generation.
Extensive experiments demonstrate that \textsc{CoNT} greatly improve the conventional MLE  baselines and significantly outperforms all previous contrastive generation models. \textsc{CoNT} establishes new state-of-the-art results on summarization, code comment generation (without external data), and data-to-text generation. Particularly, on data-to-text generation and commonsense generation, \textsc{CoNT}  achieves on-par performance with the powerful large pre-trained models: T5-large, T5-3B~\cite{raffel2019exploring} with only the base version of T5 while maintaining the  efficiency of lightweight models.

\section{Background}
\subsection{Neural Conditional Text Generation} A neural sequence-to-sequence model \citep{sutskever2014seq} $\mathcal{M}=(f, g)$ generates the target sequence conditioning on a source sequence, where $f$ and $g$ denote the encoder and decoder, respectively. It is typically trained using the language model objective with the maximum likelihood estimation (MLE). Given a source sequence $\boldsymbol{x} = \{x_i\}_{i=0}^M$ and its target sequence $\boldsymbol{y} = \{y_i\}_{i=0}^N$, we minimize the following negative log likelihood (NLL) loss:
\begin{align}
    \mathcal{L}_{\text{NLL}} &= -\sum_{t=1}^N\log p_\theta (y_t|\boldsymbol{x}, \boldsymbol{y}_{<t}). 
    \label{eq:nll}
\end{align}
At training stage,  it predict the next word based on previous ground truth input $\boldsymbol{y}_{<t}\in \boldsymbol{y}$, but at inference stage, tokens of $\boldsymbol{y}_{<t}$ are predicted by itself,  this introduces the  \textit{exposure bias}.

\subsection{Naive Contrastive Learning for Text Generation}\label{sec:naive-cl}
Contrastive text generation introduces a contrastive term in addition to the original NLL loss. In Naive CL, we simply follows SimCLR~\cite{chen2020simple} and use from-batch negative samples $\mathcal{B}$ in the InfoNCE loss~\cite{van2018representation, he2020momentum}:

\begin{align}
        &\mathcal{L}_{\text{NCE}} = -\log \frac{\exp({\cos(\mathbf{z}_{\boldsymbol{x}},\mathbf{z}_{\boldsymbol{y}})/\tau})}{\sum_{\boldsymbol{y}' \in \mathcal{B}} \exp(\cos(\mathbf{z}_{\boldsymbol{x}}, \mathbf{z}_{\boldsymbol{y}'})/\tau)}, 
        % &\mathbf{z}_X = m_f(\mathbf{H}_X),\, \mathbf{z}_Y = m_f(\mathbf{H}_Y), \,  \mathbf{z}_{Y'}\, = m_f(g(\mathbf{H}_X, Y')),\\
        % &m_f(\mathbf{H}) = \text{AvgPool}(\text{ReLU}(\mathbf{W}_{m_f}\mathbf{H} + \mathrm{b})),
    \label{eq:naive-cont}
\end{align}
where $\mathbf{z}_{\boldsymbol{x}},  \mathbf{z}_{\boldsymbol{y}}, \mathbf{z}_{\boldsymbol{y}'} \in \mathbb{R}^{d}$ denote the vector representation of input $\boldsymbol{x}$, ground truth $\boldsymbol{y}$ and negative sample $\boldsymbol{y}'\in \mathcal{B}$, respectively. $\tau$ is the temperature and $\cos(\cdot, \cdot)$ defines the cosine similarity. Intuitively, the contrastive loss $\mathcal{L}_{\text{NCE}}$ seeks to learn a similarity function that drives the distance between the source sequence representation $\mathbf{z}_{\boldsymbol{x}}$ and its ground-truth target sequence representation $\mathbf{z}_{\boldsymbol{y}}$ closer.

\section{Method}
\begin{figure}
\centering
\vspace{-0.5em}
    \includegraphics[width=0.9\linewidth]{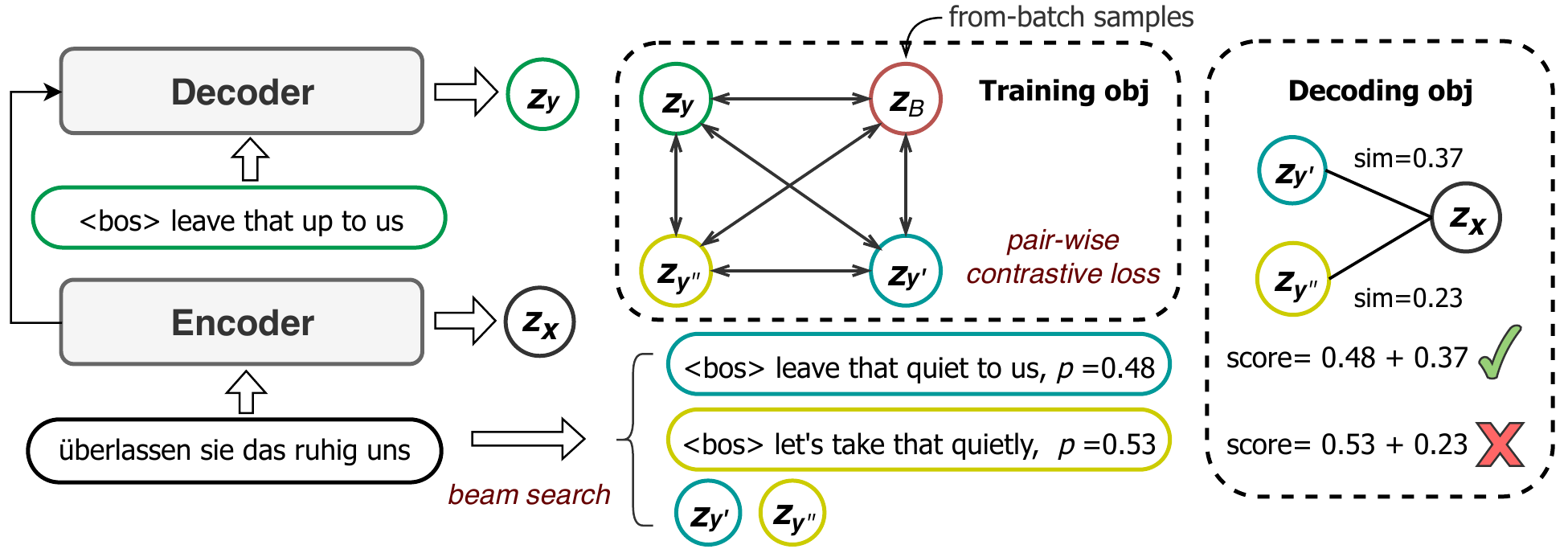}
\vspace{-0.5em}
\caption{An overview of \textsc{CoNT}. $\mathbf{z}_{\boldsymbol{x}}$, $\mathbf{z}_{\boldsymbol{y}}$ is the representation of source sequence $\boldsymbol{x}$ and its target sequence $\boldsymbol{y}$. $\boldsymbol{y}'$ and $\boldsymbol{y}''$ with their representations  $\mathbf{z}_{\boldsymbol{y}'}$, $\mathbf{z}_{\boldsymbol{y}''}$ are returned by beam search algorithm.  The feature representations come from pooling the output of the encoder (source sequence) or decoder (target sequence). 
Our training objective is obtained by comparing by all contrastive samples in pair. The decoding objective not only considers the likelihood of each sequence, but also the sequence similarity score modeled in training. 
}
\label{fig:model_overview}
\end{figure}

In this section, we present our new contrastive neural text generation framework, \method. \method advances the Naive CL (\S\ref{sec:naive-cl}) in three aspects. First, \method uses negative examples from its own predictions (\S\ref{ssec:selfgensamples}) to construct the set $\mathcal{B}$. Second, \method replaces the InfoNCE loss (Eq.\ref{eq:naive-cont}) with a N-pairs contrastive loss (Eq.\ref{eq:all_pairs_loss}) which leverages a finer-grained supervision given by the the sequence-level scores of all pairs (\S\ref{ssec:npairsloss}). Third, \method incorporates the learned similarity function into its inference score directly (\S\ref{ssec:inference}). An overview of our approach can be found in Figure~\ref{fig:model_overview}.

\subsection{Contrastive Examples from Predictions}
\label{ssec:selfgensamples}
Instead of only using contrastive examples from the same batch~\cite{chen2020simple}, we propose to add new contrastive examples from the model's own predictions.
% disturbing the ground truth in the discrete space~\cite{ssmba} or in the embedding space~\cite{lee2020contrastive} -- this is covered in related work i therefore remove this here
\citet{kalkstein2020contrast} point that using diverse contrastive samples helps  the generalization ability of the model.
Therefore, we use the diverse beam search algorithm~\cite{vijayakumar2016diverse} to create contrastive examples from the top-K list of the model's lastest predictions and then append them to the from-batch samples to form the contrastive examples. A warm-up stage where the model is only supervised by $\mathcal{L}_{\text{NLL}}$ is recommended as it guarantees the quality of the examples from the model's prediction. 
These self-generated contrastive examples alleviate the model's exposure bias. Besides, with the model's performance improving gradually, this approach creates high-quality hard negative examples that is known to be important in contrastive learning~\cite{robinson2020contrastive,kalantidis2020hard}.

\subsection{N-Pairs Contrastive Loss}
\label{ssec:npairsloss}
One major drawback of the InfoNCE loss is that it has the same treatment for  all negative samples. In text generation, this means that the relative difference between the ground truth and the contrastive examples is ignored, while this can be easily quantified using a sequence level score (e.g. BLEU) and the quality of these contrastive examples varies. To mitigate this problem, we propose to employ a pair-wise margin loss. We first rank all the contrastive examples based on an oracle function $o(\cdot, \boldsymbol{y})$, which computes a sequence-level score with the ground truth $\boldsymbol{y}$. Given a input sequence $\boldsymbol{x}$, the ground truth $\boldsymbol{y}$, and a set of $K$ contrastive samples $\mathcal{B} = {\{\boldsymbol{y}_1,\boldsymbol{y}_2,\cdots,\boldsymbol{y}_K\}}$, we can create a series of example pairs $(\boldsymbol{y}^{+}, \boldsymbol{y}^{-}) \in \mathcal{P}$, where $+$ and $-$ are determined by their ranks.\footnote{$\mathcal{P}$ contains $C_K^2$ pairs constructed from $\mathcal{B}$, ground truth $\boldsymbol{y}$, and from-batch examples.} The contrastive learning objective is formulated as a margin loss according to their cosine similarity to the source representation $\mathbf{z}_{\boldsymbol{x}}$:
\begin{equation}
\label{eq:all_pairs_loss}
\mathcal{L}_{\text{N-Pairs}} = \sum_{(\boldsymbol{y}^{+}, \boldsymbol{y}^{-}) \in \mathcal{P}} \mathcal{L}(\boldsymbol{y}^+,\boldsymbol{y}^-) = \sum_{(\boldsymbol{y}^{+}, \boldsymbol{y}^{-}) \in \mathcal{P}} \max\{0, \text{cos}(\mathbf{z}_{\boldsymbol{x}}, \mathbf{z}_{\boldsymbol{y}^-})  -  \text{cos}(\mathbf{z}_{\boldsymbol{x}}, \mathbf{z}_{\boldsymbol{y}^+})  + \xi \}.
\end{equation}
We further set $\xi = \gamma *(\text{rank}(\boldsymbol{y}^-) - \text{rank}(\boldsymbol{y}^+))$ following \citet{zhong2020extractive} to reflect the quality difference in these pairs, where $\gamma$ is a hyperparameter controlling the the strength. Full details of the training algorithm can be found in Algorithm~\ref{alg:training}, Appendix \ref{sec:limit-appendix}.

\subsection{Inference with Learned Similarity Function}
\label{ssec:inference}
Previous inference algorithm for contrastive text generation method~\cite{lee2020contrastive} usually remains the same with non-contrastive approaches. In \textsc{CoNT}, we incorporate the similarity function learned in the N-pairs contrastive loss into the decoding stage. Despite such a inference objective can be generalized to other contrastive learning methods as long as the vector representations for source and target sequence pair exist, the design of \textsc{CoNT} can better make use of the learned similarity function (\S\ref{sec:ablation_alpha}).
The decoding objective in CoNT is to find the sequence $\boldsymbol{y}^*$ that maximizes both the learned similarity score and the conventional language model likelihood:
\begin{equation}
\label{eq:infer}
    \boldsymbol{y}^* = \arg\max\limits_{\hat{\boldsymbol{y}}} \{ \alpha\cdot\cos(\mathbf{z}_{\boldsymbol{x}},\mathbf{z}_{\hat{\boldsymbol{y}}}) + 
    (1-\alpha)\prod_{t=0}^{n}\, p(\hat{y}_t|\boldsymbol{x}, \hat{\boldsymbol{y}}_{<t})\},
\end{equation}
where $\mathbf{z}_{\boldsymbol{x}}, \mathbf{z}_{\hat{\boldsymbol{y}}}\in \mathbb{R}^d$ is the vector representation of $\boldsymbol{x}$, $\hat{y}$, and  $\alpha$ is the hyperparameter that balances the contribution of each term. In most cases, $\alpha$ can be directly set to 0.5, tuning $\alpha$ on the validation set will usually get better results. Algorithm ~\ref{alg:inference} illustrates the inference stage in CoNT in details. 

\begin{figure}[h!]
\centering
    \includegraphics[width=0.95\linewidth]{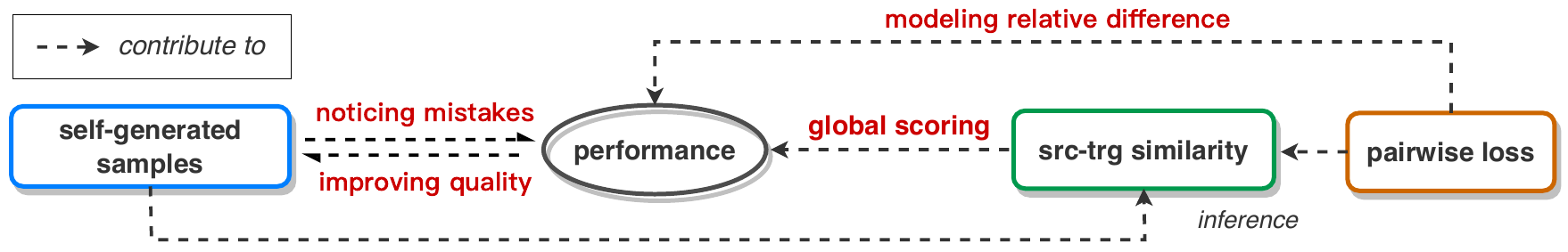}
\vspace{-0.5em}
\caption{Relationship between different modules in \textsc{CoNT}. Both the design of the pairwise loss function and self-generated samples could contribute the source-target similarity function that computes the sequence-level score at inference stage.  With the performance improved, self-generated contrastive samples tend to be more indistinguishable.}
\label{fig:model_summarize}
\end{figure}

The relationship between different modules of \textsc{CoNT} is summarized in Figure~\ref{fig:model_summarize}.

\begin{algorithm}[t!]
\caption{Inference algorithm: Given an input sequence $\boldsymbol{x}$, a contrastive generation model $\mathcal{\hat{M}} = (\hat{f},\hat{g})$; return the output sequence.}
\label{alg:inference}

\begin{algorithmic}[1]
\Procedure{BeamSearch}{$g$, $H_{\boldsymbol{x}}$, $b$}
    \algorithmiccomment{beam search algorithm}
    \State \textbf{return} Text, likelihood, logits of the $b$ hypotheses
\EndProcedure
\end{algorithmic}

\begin{algorithmic}[1]
\Procedure{Inference}{$\hat{G}$, $\boldsymbol{x}$}
    % \State $\mathcal{\hat{M}}$.no\_grad()
    % \algorithmiccomment{Do not need gradient}
    \State $\mathbf{H}_{\boldsymbol{x}} \leftarrow \hat{f}\,(\boldsymbol{x})$,  $b\leftarrow$ beam size, $\alpha \leftarrow$ balance factor $\in (0,1)$
    \State $\boldsymbol{y}^{1:b}, \mathbf{P}_{\boldsymbol{y}}^{1:b}, \mathbf{H}_{\boldsymbol{y}}^{1:b}$ = \textsc{BeamSearch$(\hat{g}, \mathbf{H}_{\boldsymbol{x}}, b)$}
    \algorithmiccomment{Get $b$ candidates with beam search}
    \State $\mathbf{z}_{\boldsymbol{x}}, \mathbf{z}_{\boldsymbol{y}}^{1:b} \leftarrow \text{Avg}(\mathbf{H}_{\boldsymbol{x}}), \text{Avg}(\mathbf{H}_{\boldsymbol{y}}^{1:b})$
    \algorithmiccomment{$\text{Avg}(\cdot)$ is an average pooling function}
    \State $\mathbf{D}_{\boldsymbol{y}}^{1:b} \leftarrow$ Cosine distance between $\mathbf{z}_{\boldsymbol{x}}$ and representation of hypotheses $\mathbf{z}_{\boldsymbol{y}}^{1:b}$
    \State $\mathbf{P}_{\boldsymbol{y}}^{1:b} \leftarrow $ Likelihood of hypotheses returned by beam search
    \State $k = \arg \max_{i=1..b} \{\alpha*\mathbf{D}_{\boldsymbol{y}}^i + (1-\alpha)*\mathbf{P}_{\boldsymbol{y}}^i\}$
    \State \textbf{return} ${\boldsymbol{y}}^{k}$
\EndProcedure
\end{algorithmic}
\end{algorithm}

\section{Experiments}
We experiment \textsc{CoNT} on 5 downstream tasks with 10 different benchmarks. Our contrastive learning framework supports most sequence-to-sequence  models at multiple scales.
Concretely, we experiment \method on 4 kinds of base models: (a) Transformer-samll (60M)~\cite{vaswani2017attention} and transformer-base (220M), (b) T5-small (60M) and T5-base (220M)~\cite{raffel2019exploring}, (c) CodeT5-base (220M)~\cite{wang2021codet5} , (d) PEGASUS-large (560M)~\cite{zhang2020PEGASUS}. Details of our experimental setup for each benchmarks can be found in Appendix~\ref{sec:exp-setting}.

% \subsection{Overview}
On WMT'16 Ro-En (machine translation) and XSum (summarization) which are also used in previous contrastive text generation frameworks~\cite{lee2020contrastive}, results show that \textsc{CoNT} is able to substantially improve the MLE baseline and clearly outperform all previous contrastive baselines by a large margin. We also build \textsc{CoNT} on state-of-the-art (SOTA) baselines: PEGASUS-large (summarization), CodeT5-base (code comment generation) and achieve new SOTA. Moreover, on data-to-text generation and commonsense generation, \textsc{CoNT} also shows its superior performance over strong MLE  baselines. 

\subsection{Baselines}~\label{sec:baseline}
\vspace{-2em}
\begin{enumerate}[itemsep=0.8mm, parsep=0pt, leftmargin=*]
 \item \textbf{Na\"ive CL}~\cite{chen2020simple}: Naive CL denotes the naive contrastive learning approach that treats the ground truth as the positive sample and the target sequences from the same mini-batch as the negative examples. The training object of naive CL takes the form of Eq.~\ref{eq:naive-cont}. We also implement Naive CL with N-Pairs contrastive loss, it can be viewed as an ablation study  when setting beam size of \textsc{CoNT} to 0 during training. 

 \item \textbf{SSMBA CL~\cite{ssmba}}: Compared with naive CL, SSMBA  builds more positive samples via disturbing the ground truth in the discrete space. Concretely, SSMBA first randomly masks 25\% tokens in the target sequence and  then reconstructs the ground truth with a masked language model BERT.  

 \item \textbf{Dropout CL~\cite{gao2021simcse}}:  Dropout CL enhances the positive samples by using dropout mechanism on the target sequence. We use the default dropout rate of standard transformer decoder~\cite{vaswani2017attention} and input the ground truth to the decoder twice.
 \item \textbf{CLAPS~\cite{lee2020contrastive}}: CLAPS is  previous the best contrastive learning framework for conditional text generation task.
 In order to  provide more challenging contrastive examples,
 CLAPS propose to simultaneously create extra positive and negative pairs by adding perturbations to the ground truth sequence in the continuous embedding space.  
 \item \textbf{\textsc{CoNT}} (this work): \textsc{CoNT} is the contrastive neural text generation framework proposed in this work. We  implement its InfoNCE version by treating ground truth as positive sample and self-generated samples are also treated as negative samples.
\end{enumerate}

% \subsection{summary}
% \lpk{give an overall comment on how CoNT perform in different tasks in general, before going to each task.}

\subsection{Quantitative Results}
\paragraph{Machine Translation}\label{sec:mt}
For machine translation, we evaluate \method on WMT 2016 Romanian-to-English translation task (WMT'16 Ro-En), WMT 2014 English-to-German translation task (WMT'14 En-De) and IWSLT 2014 German-to-English translation task (IWSLT'14 De-En).  
We use BLEU as the evaluation metric.  Results in Table~\ref{tab:nmt-resutls}  (rows with gray background)  indicates our model \textsc{CoNT} significantly improves the traditional maximum likelihood estimation training and inference framework. On WMT'16 Ro-En,  \textsc{CoNT}  outperforms previous the best contrastive learning approach CLAPS by 1.50 BLEU and exceeds the MLE baseline by \textbf{2.70} BLEU with the same base model T5-small.
We also compare the infoNCE loss used in previous methods with the N-Pairs margin loss described in Eq.~\ref{eq:all_pairs_loss}. 
Results show that the N-pairs contrasting samples generally works better than dividing all samples into predefined positive-negative categories. 
% The performance gap between CoNT and CoNT \textit{w/o sim} in Table~\ref{tab:nmt-resutls} indicates the effectiveness of the contrastive learning with target-source similarity. 
Similar to CLAPS and Naive CL, only incorporating contrastive learning into training improves the performance of T5-small baseline on WMT'16 Ro-En to 30.55 (+\textbf{2.34}) BLEU. 
If we further add learned target-source similarity as decoding target as Eq.~\ref{eq:infer}, the result is further boosted to 30.91 BLEU. 
We observe that the benefits of introducing sequence similarity into inference is more obvious on IWSLT'14 De-En -- the additional decoding target improves the vanilla beam search algorithm up to 0.86 BLEU.

\renewcommand\arraystretch{1.0}
\begin{table}[t]
\begin{center}
\caption{BLEU on WMT'16 Ro-En, IWSLT'14 De-En and WMT'14 En-De   translation tasks. For IWSLT'14 De-En and WMT'14 En-De, we use  Transformer-small (\textbf{Tr-small}) and Transformer-base (\textbf{Tr-base}) as baselines. For WMT'16 Ro-En, we add a pre-trained baseline \textbf{T5-small}.   \textit{w/o seq sim} means we use the origin beam search without target-source representation similarity. The best results in each block are underlined and the best results are in bold. Rows in gray denotes the contrastive learning based model strongly outperforms its MLE version. Results with $^\dag$ are token from~\cite{lee2020contrastive}. 
}
\label{tab:nmt-resutls}
%\scalebox{1.0}{
\tabcolsep0.04 in
\begin{tabular}{lccccccc}
\toprule
\multirow{2}{*}{\vspace{-2mm}\bf Model} & \multicolumn{2}{c}{\bf WMT'16 Ro-En} & & \multicolumn{1}{c}{\bf IWSLT'14 De-En} & &\multicolumn{1}{c}{\bf WMT'14 En-De}  \\
\cmidrule{2-3} \cmidrule{5-5} \cmidrule{7-7} 
& \bf Tr-small & \bf T5-small & &\bf Tr-small & &\bf Tr-base \\ 
\midrule
MLE & 25.78 & 28.21 & &34.18 & & 27.30\\
\midrule
\multicolumn{7}{c}{\textit{Contrastive loss: InfoNCE loss}} \\
\midrule
Naive CL  & 25.49 & 27.79 & & 34.45 & & 27.28\\
SSMBA CL  & 25.98 & 28.48 & & 34.32 & & 27.16\\
Dropout CL  & \underline{26.01} & 29.10 & & 34.41 & & 27.34\\
CLAPS$^\dag$   & 23.59 & 29.41 & & -- & & -- \\
\textsc{CoNT}  & 25.74 & \underline{29.64} & & \underline{34.46} & & \underline{27.35}\\
\midrule
\multicolumn{7}{c}{\textit{Contrastive loss: N-Pairs loss}} \\
\midrule
Naive CL  & 26.15 & 29.86 & & 34.47 & & 27.41\\
\quad \textit{w/o seq sim}   & 26.27 & 29.74 & & 34.26 & & 27.45\\
\rowcolor{gray!20}
\textsc{CoNT} & {\textbf{27.70}} & {\textbf{30.91}} & & {\textbf{35.55}} & & {\textbf{28.04}}\\
\quad \textit{w/o seq sim}   & 27.42 & 30.54 & & 34.69  & & 27.77\\
\bottomrule
\end{tabular}
\end{center}
\end{table}

\paragraph{Summarization}\label{sec:summ}
For summarization, we use the XSum~\cite{narayan2018don} dataset collected from BBC News whose reference summaries are provided by human writers. We also evaluate \textsc{CoNT} on a multi-document summarzation dataset multi-news~\cite{fabbri2019multi} consisting of news articles from the site newser.com. Compared with the common summarization task, multi-document is  more challenging  where the model need to automatically summarize several articles and usually has to handle long input sequence and target sequence.  Experimental results are in Table~\ref{tab:summary-results}. The first block includes the performance of different contrastive frameworks with T5-small. On XSum, it shows that our proposed model  strongly outperform previous contrastive frameworks by about 2.0 ROUGE-1 score. We also illustrate our method is not restricted to the small model. By employing \textsc{CoNT} on  state-of-the-art base model PEGASUS, it is able to establish new state-of-the-art on the two summarization benchmarks.

\renewcommand\arraystretch{1.1}
\begin{table}[t]
\begin{center}
\vspace{-1em}
\caption{ROUGE score on Summarization datasets. Results with $^\dag$ are token from~\cite{lee2020contrastive} and results with $^*$ are from~\cite{zhang2020PEGASUS}. Current state-of-the-art models and the best results are in bold.  Previous SOTA means the best results before CoNT. }
\label{tab:summary-results}
%\scalebox{1.0}{
\tabcolsep0.09 in
\begin{tabular}{lcccccccc}
\toprule
\multirow{2}{*}{\vspace{-2mm}\bf Model} & \multicolumn{3}{c}{\bf XSum} & & \multicolumn{3}{c}{\bf Multi-News} \\
\cmidrule{2-4} \cmidrule{6-8} 
& \bf R-1 & \bf R-2 &\bf  R-L & &  \bf R-1 & \bf R-2 &\bf  R-L \\ 
\midrule
T5-small & 36.10 & 14.72 & 29.16 & & 42.36 & 15.34 & 21.91\\
T5-SSMBA CL  & 36.58 & 14.81 & 29.68 & &42.06 & 14.98 & 21.73\\
T5-Dropout CL  & 36.82 & 14.93 & 29.26 & &42.43 & 15.32 & 21.95\\
T5-CLAPS$^\dag$   & 37.89 & 15.78 & 30.59 & & -- & -- &  -- \\
T5-Naive CL & 36.34 & 14.81 & 29.41 & &42.20 & 15.18 & 21.78\\
T5-Naive CL (N-Pairs)& 37.76 & 15.48 & 30.15 & & 43.04 & 15.83 & 22.03\\
\rowcolor{gray!20}
T5-\textsc{CoNT} & \underline{39.66} & \underline{16.96} & \underline{31.86} & &\underline{44.08} & \underline{16.39} & \underline{22.58}\\
\midrule
Previous SOTA$^*$  & 47.61 &24.57 & 39.44 & & 47.52 &18.72 & 24.91\\
PEGASUS (base)$^*$ & 39.79 & 16.58 & 31.70 & & 42.24 & 13.27 & 21.44\\
\bf PEGASUS (large)$^*$ & 47.21 & 24.56 & 39.25 & & 47.52 & 18.72 & {\textbf{24.91}}\\
\rowcolor{gray!20}
PEGA-\textsc{CoNT} & {\textbf{47.76}} & {\textbf{24.69}} & 
{\textbf{39.46}} & & {\textbf{48.68}} & {\textbf{19.29}} &24.58 \\
\bottomrule
\end{tabular}
\end{center}
\end{table}

\paragraph{Code Comment Generation}\label{sec:code}
\begin{figure}[!t]
\begin{minipage}[t]{0.48\textwidth}
\centering
\captionof{table}{{BLEU on two code comment generation datasets Java and Python. Results with $^\dag$ and $^*$ are from~\cite{wang2021codet5}}
% Superscript $^\star$ denotes our reproduced numbers, others are reported directly from the respective papers.
}
\label{tab:code-results}
\resizebox{1.0\columnwidth}{!}{%
\setlength\tabcolsep{8pt}
\begin{tabular}{lcc}
\toprule
{\bf Model}   & {\bf Python} & \textbf{Java}\\

\midrule
CodeBERT $^\dagger$ &19.06 & 17.65 \\
PLBART $^\dagger$     & 19.30 & 18.45 \\
CodeT5 $^\dag$ & 20.01 & 20.31 \\
\bf CodeT5-Dual-Gen$^\dag$ & \underline{20.11} & \underline{20.41} \\
\midrule
\multicolumn{3}{l}{\bf With N-Pairs CL }\\
CodeT5-Naive CL &20.26  & 20.31 \\
CodeT5-\textsc{CoNT} & \underline{20.43} &  \underline{20.56}\\
\midrule
\multicolumn{3}{l}{\bf With External Training Data }\\
CodeT5-Multi-Task$^\dag$  & 20.36 & 20.46\\
\bf REDCODER$^*$  & \textbf{21.01} & \textbf{22.94}\\
\bottomrule
\end{tabular}}
\vspace{-1em}
% \label{table:data_diverse_wmt}
\end{minipage}
\hfill
\begin{minipage}[t]{0.48\textwidth}
\centering
\captionof{table}{BLEU on data-to-text generation dataset WikiBio. We run our model three times and report  the mean and variance of the BLEU metric. Results with $^\dag$ are token from \cite{liu2018table} and results with $^*$ are from~\cite{arbabi2021r2d2}.}
\label{tab:wikibio-results}
\resizebox{0.9\columnwidth}{!}{%
\setlength\tabcolsep{9pt}
\begin{tabular}{lc}
\toprule
{\bf Model}   & {\bf BLEU}\\
\midrule
Table NLM $^\dagger$ & 34.70 \small{$\pm$0.36}  \\
vanilla Seq2Seq $^\dagger$ & 42.06 \small{$\pm$0.32}  \\
StructureAware $^\dagger$ & 44.89\small{$\pm$0.33}  \\
\bf R2D2 $^*$     & \underline{46.23}\small{$\pm$0.15} \\
T5-small $^\dagger$     & 46.02\small{$\pm$0.36}  \\
\midrule
\multicolumn{2}{l}{\bf With N-Pairs CL }\\
T5-small-Naive CL & 46.50\small{$\pm$0.24}  \\
\rowcolor{gray!20}
T5-small-\textsc{CoNT} & \textbf{47.17}\small{$\pm$0.19}  \\
\bottomrule
\end{tabular}}
% \label{table:data_diverse_wmt}
\end{minipage}
\end{figure}

Code comment generation aims to generate an English description for a function-level code snippet. We test our method on two widely used datasets Java and Python from the CodeXGLUE benchmark~\cite{lu2021codexglue}.  Results are shown in Table~\ref{tab:code-results}. Our model is built upon state-of-the-art pre-trained model on program language model CodeT5-base. CodeT5-Dual-Gen means they further involve a comment-to-code task which is the best model on Python and Java without using external data.  We also include the results of earlier strong pre-trained baselines: PLBART  and CodeBERT. We also report some data augmentation methods in the third block of Table~\ref{tab:code-results}. CodeT5-Multi-Task makes use of training datasets of other program languages. REDCODER~\cite{parvez2021retrieval}  uses retrieval to enhance the task with open-source code base and achieve the bset results on this task. Our model is orthogonal to these methods and  clearly outperforms all baselines without external data.

\paragraph{Data-to-text Generation}\label{sec:datatotext}
Data-to-text generation aims to produce text from non-linguistic input. The first benchmark we use is WikiBio~\cite{lebret2016neural} consisting of  biography pairs from English Wikipedia where the infobox is treated as input sequence, and the target sequence is the first sentence of the biography. Totto~\cite{parikh2020totto} is also collected from Wikipedia whose input is a table with its highlighted cells and  target sequences are professionally annotated by human. Results on WikiBio is shown in Table~\ref{tab:wikibio-results}, the performance of some popular baselines (first block) are token from~\cite{liu2018table}. R2D2~\cite{arbabi2021r2d2}, using  XLNet~\cite{yang2019xlnet}-large as base model, is previous state-of-the-art model on WikiBio. We experiment  \textsc{CoNT} on T5-small, and results show that we exceed  R2D2 by about 0.94 BLEU.  
The test set and dev set of Totto are both split into
two parts - overlap and non-overlap. The non-overlap part contains out-of-domain samples from the training set.
The test set of Totto is invisible and we report the results on dev set and test set by the feedback of the Totto authors.
We also add PARENT~\cite{dhingra2019handling}  and BLEURT~\cite{sellam2020bleurt} as evaluation metrics. PARENT is a word-overlap based metric that designed to evaluate the factual accuracy of generation results. BLEURT is trained under human supervision and correlates well with human judgement. As we can see from Table~\ref{tab:totto-results}, by comparing our model with  different T5 variants, we show that \textsc{CoNT} is able to greatly outperform the large version of T5 even built on a T5-base model. Compared with previous state-of-the-art model T5-3B, our model still show advantage in PARENT and BLEURT.

\renewcommand\arraystretch{1.1}
\begin{table}[t]
\begin{center}
\caption{Results on the dev set and test set Totto. PAR is short for PARENT score. Dev Set (Non) means the non-overlap subset of the dev set. results with $^\dag$ are reported in~\cite{kale2020text}.}
\label{tab:totto-results}
\tabcolsep0.04 in
\begin{tabular}{lcccccccccc}
\toprule
\multirow{2}{*}{\vspace{-2mm}\bf Model} & \multicolumn{2}{c}{\bf Dev Set (All)} & &\multicolumn{2}{c}{\bf Dev Set (Non)} & & \multicolumn{3}{c}{\bf Test Set (All)} \\
\cmidrule{2-3} \cmidrule{5-6}    \cmidrule{8-10}
& \bf BLEU & \bf PAR &  &\bf BLEU  &\bf PAR & &  \bf BLEU & \bf PAR  &\bf BLEURT \\ 
\midrule
BERT-to-BERT$^\dag$ &  44.0 & 52.6 & & 34.8 & 46.7 & & 44.0 & 52.6 & 0.121\\
T5-large$^\dag$  & 48.1 & 57.3 & & 39.8 & 52.8 & & -- & -- & --\\
\bf T5-3B$^\dag$ & {48.4} & {57.8}&  & {40.4} & {53.3} & & \bf49.5 & {58.4} & {0.230} \\
T5-base$^\dag$ & 47.7 & 57.1 & & 39.6 & 52.6 & & -- & -- & --\\
\rowcolor{gray!20}
T5-base-\textsc{CoNT} &  \bf49.2 &  \bf59.4 &  & \bf41.5  & \bf55.0 & & {49.1} & \bf58.9 & \bf0.238\\
\bottomrule
\end{tabular}
\end{center}
\end{table}

\paragraph{Commonsense Generation}\label{sec:commonsense}
The task of commonsense generation  aims to explicitly test the ability
of machines on commonsense reasoning. The source sequence consists of a set of concepts and the target sequence is a fluent sentence mentioning all the input concepts.  We evaluate \textsc{CoNT} on the CommonGen~\cite{lin2020commongen} benchmark with a hidden test set and the final results is obtained with the help of the authors of CommonGen. In addition to the mostly used metrics, CIDEr~\cite{vedantam2015cider} and SPICE~\cite{anderson2016spice}, concerning to evaluating semantic faithfulness, are highlighted by the leaderboard of commonGen. In Table~\ref{tab:commonGen-results}, we demonstrate that the lightweight T5-base model is able to greatly benefit from our contrastive learning framework.  Moreover, \textsc{CoNT} not only surprisingly outperforms its MLE baseline but also surpass the large version of T5 in terms of CIDEr and SPICE metrics.

\renewcommand\arraystretch{1.1}
\begin{table}[t]
\begin{center}
\vspace{-1em}
\caption{Results on CommonGen. Results with $^\dag$ are reported in~\cite{lin2020commongen}.  The metrics used in the official leaderboard are in bold. Human performance is also reported as an upper bound.}
\label{tab:commonGen-results}
\tabcolsep0.04 in
\begin{tabular}{lccccccccccc}
\toprule
{\bf Model} & \multicolumn{2}{c}{ROUGE-2/L} & &\multicolumn{2}{c}{BLEU-3/\bf4} & METEOR & \bf CIDEr &\bf SPICE & Coverage \\
\midrule
GPT-2$^\dag$ & 16.85 & 39.01 & & 33.92 & 23.73 & 26.83 & 12.19 & 23.57 & 79.09\\
% UniLM$^\dag$  & 21.20 & \bf43.60 & & 41.82 & 30.73 &30.62 & 14.89 & 27.43 & 89.19 \\
BART$^\dag$  & \bf22.02 & 41.78 & & 39.52 & 29.01 & 31.83 & 13.98 & 28.00 & \bf97.35 \\
T5-large$^\dag$ & 21.74 & 42.75 & & \bf43.01 & \bf31.96 & 31.12 & 15.13 & 28.86 & 92.29\\
T5-base$^\dag$ & 14.63 & 34.56 & & 28.76 & 18.54 & 23.94 & 9.40  & 19.87 &  76.67 \\
\rowcolor{gray!20}
T5-base-\textsc{CoNT} & 20.96 & \textbf{43.15} & & 42.60 & 31.42 & \bf32.05& \bf15.96 & \bf28.95& 96.55\\
\midrule
Human  & 36.72 & 54.45 & & 52.55 & 46.49 & 38.79 & 37.64 & 52.43 & 99.33 \\
\bottomrule
\end{tabular}
\end{center}
\end{table}

\paragraph{Advanced Evaluation Metrics}
Considering the training efficiency of \textsc{CoNT}, we mainly select the lexical matching metrics as oracle measurement function. To verify that the improvement brought by \textsc{CoNT} is not due to the over-fitting of lexical matching metrics, we further evaluate generated text with  advanced metrics based on neural models: BERTScore~\cite{zhang2019bertscore} and BLEURT~\cite{sellam2020bleurt}. For BERTScore, we use their roberta-large\_L17\_no-idf\_version and for BLEURT we use the default setting provided on their github\footnote{https://github.com/google-research/bleurt}. Results are shown in Table~\ref{tab:adv-results}. The base model used on IWSLT'14 De-En is transformer small and on the other datasets we select T5 as the base model. For all datasets \textsc{CoNT} also make non-trivial improvements  in terms of the two neural metrics. Particularly, \textsc{CoNT} improve the results of MLE model on IWSLT'14 De-En by 0.03 BLEURT and improve the results of MLE model on XSum by 2.68  BERTScore. 

To get more accurate and convincing results, we also conduct a ranking based human evaluation  on two mainstream tasks: machine translation (IWSLT'14 De-En) and text summarization (XSum) with 60 samples for each tasks. Following~\citet{cheng2016neural}, we hired 2 annotators asking them to rank the given candidate output based on fluency, coherence, and their personal preference ( rank these systems 1st, 2nd, and so on) and we calculate the average ranking. For each sample, there are four candidates consist of a human-written reference, a sequence from MLE model, a sequence from Naive CL, and a sequence from \textsc{CoNT}.  Table~\ref{tab:human-results} shows the results of our human evaluation. Generally \textsc{CoNT} outperform all baseline systems according to the average ranking.

\renewcommand\arraystretch{1.1}
\begin{table}[t]
\begin{center}
\caption{BLEURT and BERTScore on the test set of 4 translation and summarization datasets. The first column of each dataset represents BLURT and the second column is BERTScore. }
\label{tab:adv-results}
\tabcolsep0.04 in
\begin{tabular}{lccccccccccccc}
\toprule
\multicolumn{1}{l}{\bf Model} & & \multicolumn{2}{c}{\bf IWSLT'14 De-En} & &\multicolumn{2}{c}{\bf WMT'16 Ro-En} & & \multicolumn{2}{c}{\bf XSum} & & \multicolumn{2}{c}{\bf Multi-News}  \\
% \midrule
\cmidrule{1-1} \cmidrule{3-4} \cmidrule{6-7} \cmidrule{9-10} \cmidrule{12-13}
MLE  & \quad & 0.137 & 62.28 & & 0.272 & 69.09 & & -0.552 & 44.10 & & -0.568 & 17.21 \\
\textsc{CoNT} & \quad & 0.167 & 63.38 & & 0.281 & 69.33 & & -0.462 & 46.78 & & -0.505 & 17.47 \\
\bottomrule
\end{tabular}
\end{center}
\end{table}

\renewcommand\arraystretch{1.1}
\begin{table}[t]
\begin{center}
\caption{Results of human evaluation on the test sets of  translation and summarization.}
\label{tab:human-results}
\tabcolsep0.04 in
\begin{tabular}{lcccccccccccc}
\toprule
\multirow{2}{*}{\vspace{-2mm}\bf Model} & \multicolumn{5}{c}{\bf Machine translation} & & \multicolumn{5}{c}{\bf Summarization} \\
\cmidrule{2-6}    \cmidrule{8-12}
& 1st & 2nd & 3rd & 4th &\bf avg rank & & 1st & 2nd & 3rd & 4th & \bf avg rank\\ 
\midrule
 Ground truth& 0.88 & 0.08 & 0.0 & 0.04 & 1.2 & & 0.52 & 0.25 & 0.12 & 0.12 & 1.86 \\
\textsc{CoNT}& 0.07 & 0.5 & 0.31 & 0.12 & 2.48 & & 0.27 & 0.4 & 0.2 & 0.13 & 2.2 \\
 Naive CL   & 0.02 & 0.25 & 0.4 & 0.33 & 3.04 & & 0.12 & 0.22 & 0.38 & 0.28 & 2.82 \\
 MLE        & 0.03 & 0.17 & 0.28 & 0.52 &3.29 & & 0.1 & 0.13 &0.3 & 0.47 & 3.13 \\

\bottomrule
\end{tabular}
\end{center}
\end{table}

\subsection{Discussion}
\begin{figure}
\vspace{-0.3em}
\centering
\begin{subfigure}{.32\textwidth}
    \begin{center}
        \includegraphics[width=\textwidth]{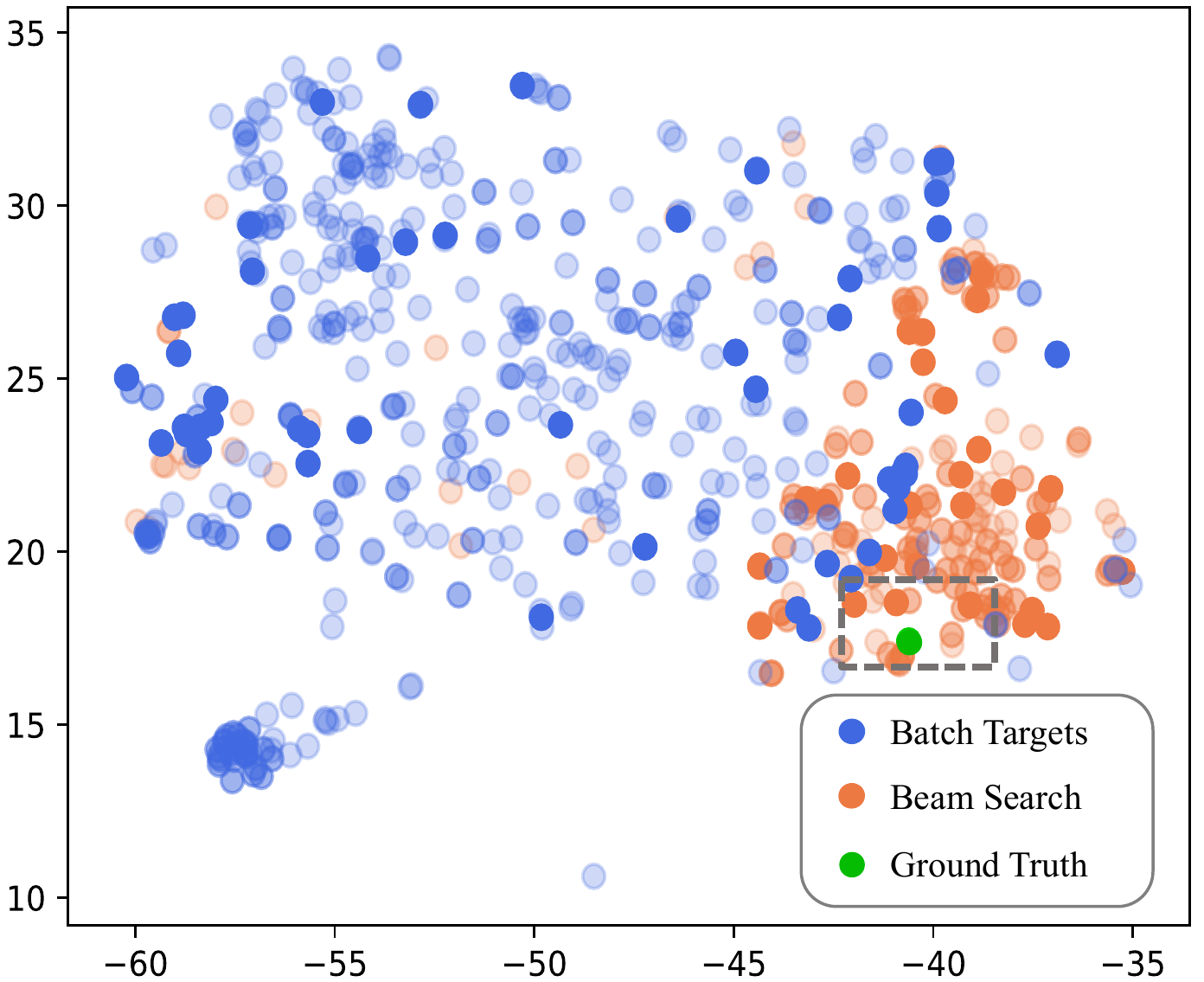}
        \caption{MLE}
        \label{fig:tsne-mle}
    \end{center}
\end{subfigure}
\hspace{1mm}
\begin{subfigure}{.32\textwidth}
  \begin{center}
    \includegraphics[width=\textwidth]{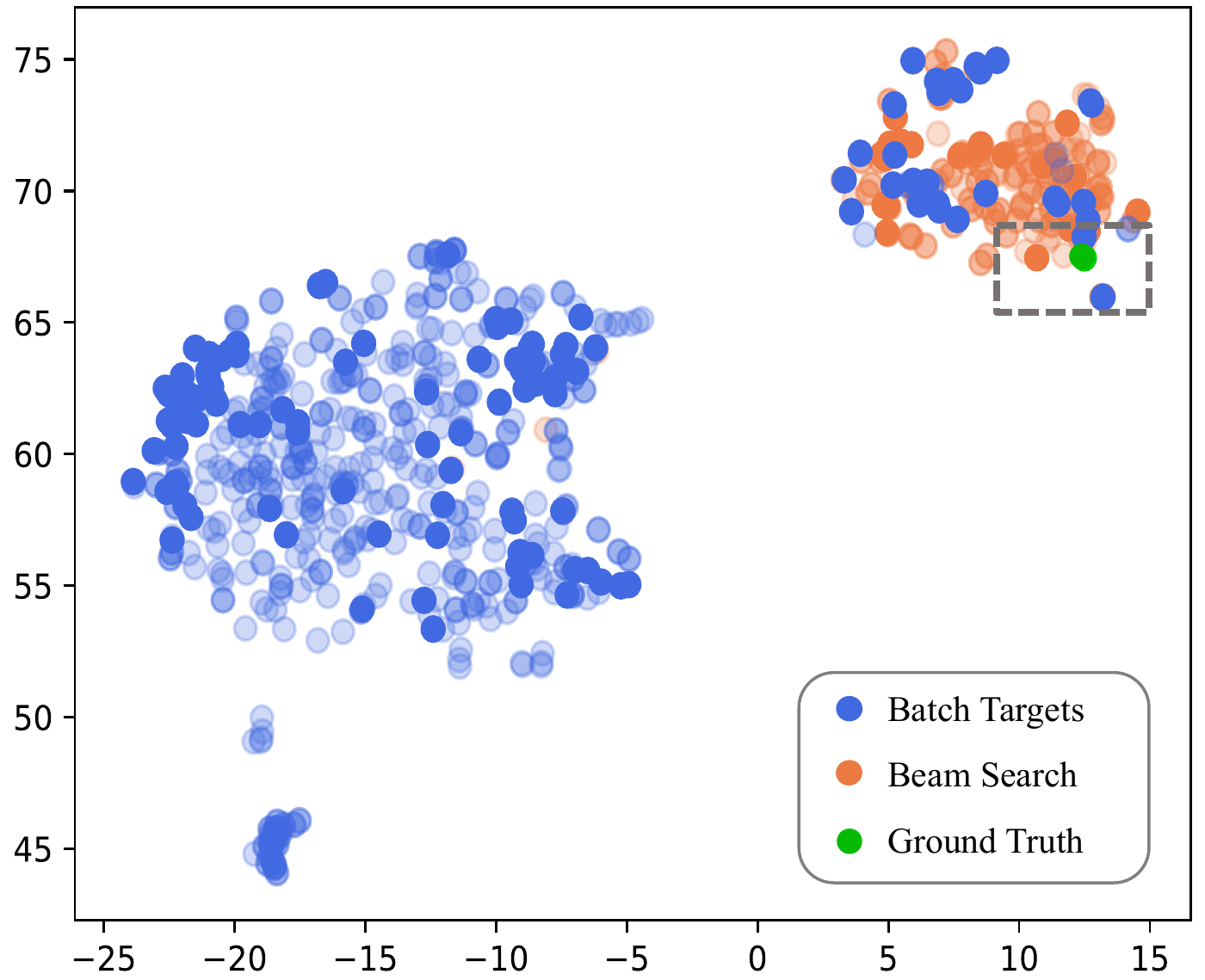}
    \caption{Naive CL}
    \label{fig:tsne-naive}
  \end{center}
\end{subfigure}
\hspace{1mm}
\begin{subfigure}{.32\textwidth}
  \begin{center}
    \includegraphics[width=\textwidth]{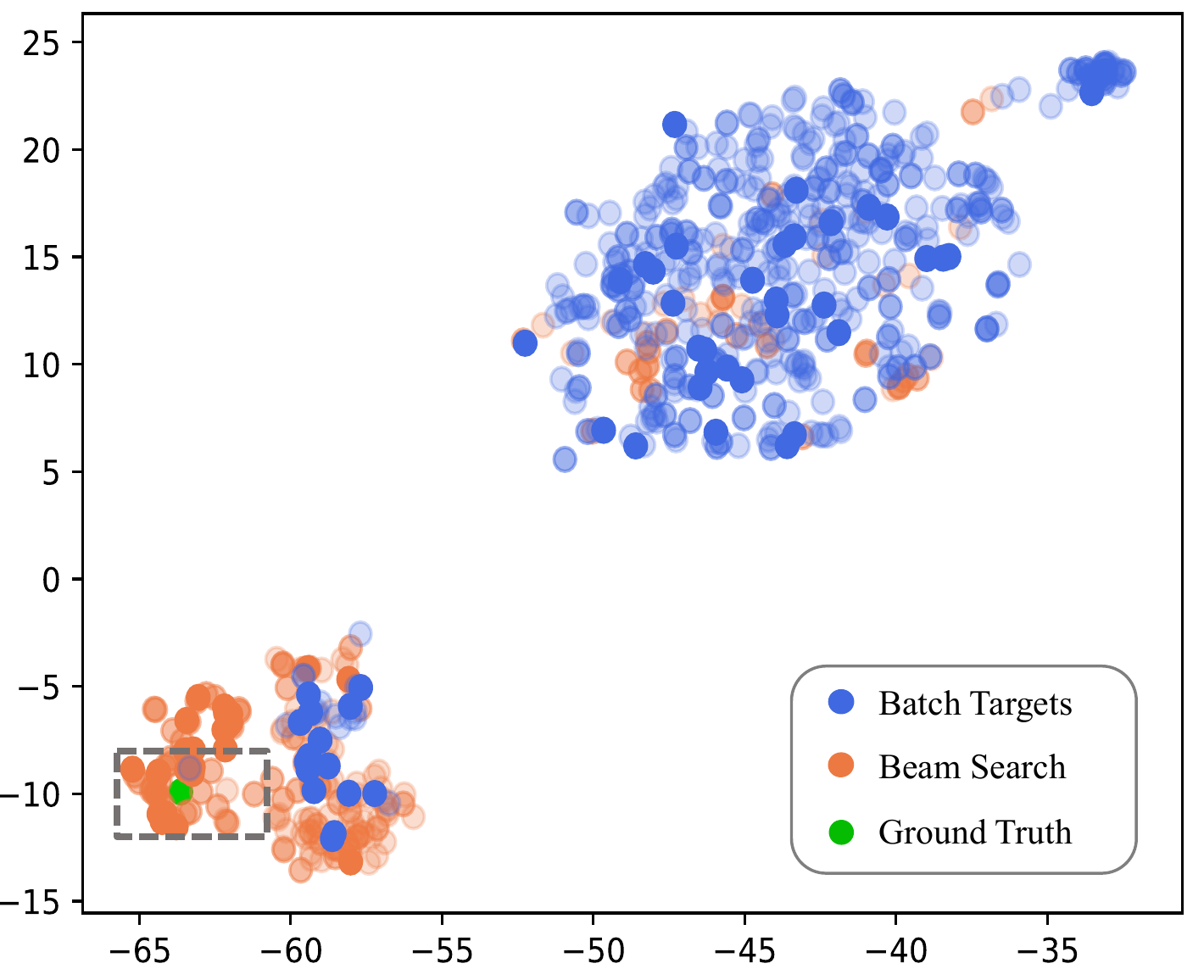}
    \caption{\textsc{CoNT}}
    \label{fig:tsne-cont}
  \end{center}
\end{subfigure}
\vspace{-0.5em}
\caption{T-SNE experiments on IWSLT'14 De-En. Each point represents a target sequence. Batch target is blue; beam search hypothesis is orange; ground-truth sequence is green. Darker points indicate sequences with higher BLEU.}
\label{fig:tsne-exp}
\end{figure}

\paragraph{Discrimination of Hard Negative Samples}
To deeply look into the learnt representations, we visualize target sequences, that is trained by MLE, naive CL and \method on IWSLT'14 De-En, with the t-SNE algorithm~\cite{van2008visualizing}.
The visualized sequences consist of three groups of target sequence: a) batch targets that is mostly unrelated to ground -truth target; b) beam search hypothesis that could be of high/low quality; c) ground truth target. As can be seen in Figure~\ref{fig:tsne-mle}, the representations trained by MLE are distributed almost uniformly in the vector space, and there are no clear boundary between one group to another.
With naive CL, we find there is clear boundary between batch tokens and others.
Naive CL does contribute to discriminating related sequences with unrelated ones, but it still cannot distinguish hypothesis of high quality from the others. 
Even without contrastive learning, the generation model trained has already pulled from-batch samples away from the ground truth and the naive contrastive learning procedure is only to make the margin more obvious.
As for \method, a set of hypotheses of low quality are excluded from the neighborhoods of the ground-truth targets.
The experimental results verifies that \method enables better representations in sequence generation.

\paragraph{Sequence Likelihood and Sequence Similarity}\label{sec:ablation_alpha}
\begin{figure}
\vspace{-0.5em}
\centering
\begin{subfigure}{.43\textwidth}
    \begin{center}
        \includegraphics[width=\textwidth]{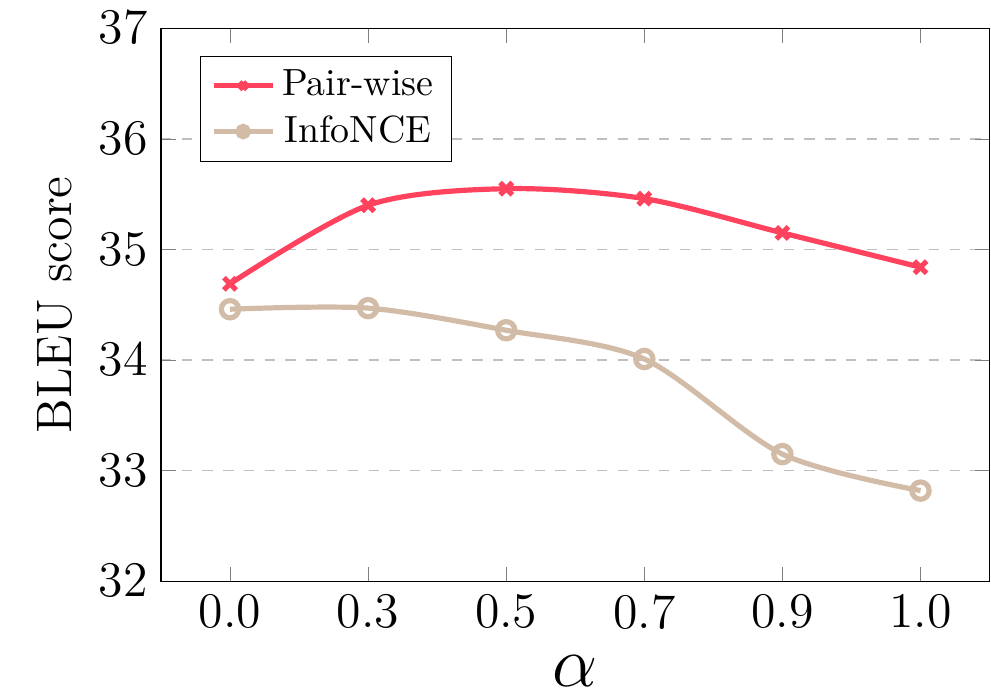}
        \caption{Relationship between $\alpha$ and BLEU with different contrastive loss. }
        \label{fig:ab-loss}
    \end{center}
\end{subfigure}
\hspace{2mm}
\begin{subfigure}{.43\textwidth}
  \begin{center}
  \vspace{0.5em}
    \includegraphics[width=\textwidth]{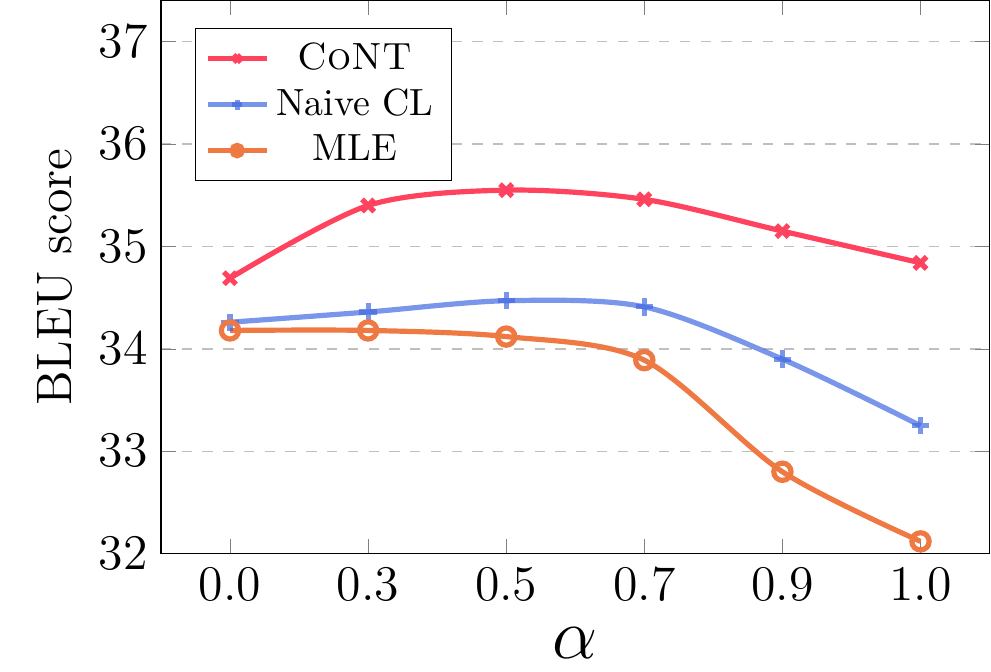}
    \caption{Relationship between $\alpha$ and BLEU  with different training methods.}
    \label{fig:ab-sample}
  \end{center}
\end{subfigure}
\vspace{-0.5em}
\caption{Ablation study on the balance factor $\alpha$ on the test set of IWSLT'14 De-En where $\alpha=0.0$ means selecting output only relying on likelihood and $\alpha=1.0$ means choosing output with only sequence similarity.}
\label{fig:ab}
\end{figure}

We also perform ablation study on our ranking objective and self-generated negative samples on  IWSLT'14 De-En with Transformer-small as  base model.
For the ranking objective, we compare our N-pairs contrastive objective with InfoNCE with increasing $\alpha$.
Figure~\ref{fig:ab-loss} that N-pairs contrastive loss consistently outperform InfoNCE.
With $\alpha \in [0.3, 0.7]$, N-pairs contrastive loss can benefit  target-source representation similarity, a sequence-level score in inference process, while InfoNCE cannot.
For the sampling strategy, we compare \method with naive CL and vanilla MLE. 
\method use self-generated hypotheses as negatives while naive CL only use samples within the same mini-batch.
Figure~\ref{fig:ab-sample} shows that contrastive learning with self-generated hypotheses is more effective than using batch samples.
With $\alpha \in [0.3, 0.7]$, \method gains about $1.0$ BLEU improvements, while naive CL is only improved with less than $0.5$ BLEU.

\section{Related Work}
\paragraph{Contrastive Learning}

Contrastive learning~\cite{schroff2015facenet, he2020momentum, chen2020simple} aims to learn a better representation via contrasting positive and negative samples.
It also has been widely used in the field of natural language processing~\cite{RDevonHjelm2018LearningDR,gao2021simcse,kong2020mutual, krishna2022rankgen, zhou-etal-2022-knn}. 
\citet{jiang2020robust} show that contrastive learning helps learn a robust pre-trained model and \citet{lee2020contrastive} first introduce contrastive learning into text generation to mitigate the exposure bias problem. They propose an adversarial method to build more challenging positive-negative samples in additional to the from-batch samples.  SimCTG~\cite{su2022contrastive} is also a contrastive framework on text generation. Our work  differs from SimCTG from  motivation and method. They introduce contrastive leanrning mainly to encourage diversity which is important in dialogue systems. And they perform token-level contrastive learning while our method focus on sequence-level contrastive examples. 

Adopting binary supervision in contrastive loss is originally proposed in FaceNet~\cite{schroff2015facenet} which helps learn the face recognition of the same person in various positions and angles. Given an anchor face image $\mathbf{x}$, a positive sample $\mathbf{x}^+$ (usually the same person) and a negative sample $\mathbf{x}^-$ from other people, the triplet loss makes $\mathbf{x}^+$ become close to $\mathbf{x}$ and maximize the distance between $\mathbf{x}$ and $\mathbf{x}^-$. 
The pair-wise contrastive loss has also been widely used in metric learning~\citet{wang2014learning,kim2019deep, chen2019energy}.
\citet{sohn2016improved} extend the triplet loss to multi-class and multi-pair.  Recent work thinks the margin value between samples should not be fixed. \citet{zhou2020ladder} divide the sample set into multiple subsets and assign different margin value to different subsets. \citet{ha2021deep,zhong2020extractive} suggests dynamically adjust the margin value via a determination metric.

\paragraph{Post-generation Re-ranking Models}
Post-generation re-ranking  re-score the multiple output sequences via training another re-ranking module.
Noisy Channel Modeling (NCM)~\cite{ng2019facebook, yee-etal-2019-simple} is a widely-used re-ranking scheme for neural machine translation. NCM parameterizes the noisy channel probability with a sequence-to-sequence model. There also various structures to instantiate the re-ranking module: \citet{gulcehre2017integrating} select the candidate with a language model, \citet{bhattacharyya2020energy} leverage an energy-based model in NMT and \citet{salazar2019masked,liu2021simcls} re-score candidates with masked language models such as BERT.  Despite this paradigm achieves impressive results while having a large size of candidate sequences, most of post-generation re-ranking systems trade efficiency and simplicity for accuracy.

\section{Conclusion}
We introduce a new contrastive neural text generation framework called \textsc{CoNT}. It models an additional contrastive learning objective to provide a sequence-level supervision for auto-regressive neural text generation models. We explore three shortcomings that limit the development of contrastive learning on text generation tasks.
Results on five generation tasks with ten different benchmarks show that \textsc{CoNT} not only clearly beats all previous contrastive generation models, but also boosts the performance of state-of-the-art large models to a new level.  \textsc{CoNT} practically does not have a negative impact on decoding speed.  Nevertheless, \textsc{CoNT} suffers from the training inefficiency problem.   In general, the total training 
time of \textsc{CoNT} is about 2$\sim$4 times more than that of a MLE based model.  A detailed discussion and some speed-accuracy trade-off tricks can be found in  Appendix~\ref{sec:limit-appendix}. Speeding up the training stage without losing accuracy is the next important step to improve \textsc{CoNT}. 

\section*{Acknowledgement}
We would like to thank the anonymous reviewers for their valuable advice. This research was supported in part by the joint research scheme of the
National Natural Science Foundation of China (NSFC) and the Research Grants Council (RGC) under grant number N HKU714/21.
\medskip
{
\small
\bibliography{neurips_2022}
}

%%%%%%%%%%%%%%%%%%%%%%%%%%%%%%%%%%%%%%%%%%%%%%%%%%%%%%%%%%%%

%%%%%%%%%%%%%%%%%%%%%%%%%%%%%%%%%%%%%%%%%%%%%%%%%%%%%%%%%%%%

\newpage
\appendix
\section{Overview}
In the supplementary materials of this work, we first discuss the main limitation of \textsc{CoNT}. After that, we describe the detailed experimental setup for the 10 different benchmarks. Finally, we randomly present some generation examples of \textsc{CoNT} on machine translation and summarization.

\section{Limitation}~\label{sec:limit-appendix}
\textsc{CoNT} practically does not have a negative impact on decoding speed. Compared with the decoding algorithm used in the inference stage of conventional generation models, the additional operations brought by~\textsc{CoNT} are reflected in line 4 and line 5, Algorithm~\ref{alg:inference}. The two operations can be efficiently calculated on GPUs.
However, an obvious disadvantage of \textsc{CoNT} is the sacrifice of training efficiency.   In general, the total training 
time of \textsc{CoNT} is about 2$\sim$4 times more than that of a MLE based model. 

We show the pseudo code of our training procedure in Algorithm~\ref{alg:training}. As can be seen from this algorithm, there are three main factors that harm the training speed of \textsc{CoNT}: (i) a pre-train stage to ensure meaningful contrastive samples (line 3 in Algorithm~\ref{alg:training}), (ii) the involvement of token-by-token decoding in training (line 7 in Algorithm~\ref{alg:training}) which can hardly benefit from the parallel computing power of GPU, and (iii) the calculation of oracle measurement (the nested loop in line 9$\sim$11 Algorithm~\ref{alg:training}). Regrading the 
last issue, if we use some lexical matching metrics such as BLEU or ROUGE that always need calculated on CPU, the training speed will  be obviously slowed down. During waiting for the results from CPU, the GPU utilization will decrease to 0. We solve the problem with by placing the nested loop on GPU where candidate samples and the ground truth are represented by a long type tensor and the similarity between two sequences are calculated by matrix multiplication which significantly improve the efficiency. The GPU version  of getting oracle measurement  will also be released along with our source code.  As for the first issue and the second issue, we haven't found good alternatives yet so that we advise two ways to make a speed-accuracy trade-off at the following parts. 

\begin{algorithm}[h!]
\caption{Contrastive Text Generation: Given a  generation dataset <$\mathcal{X}$, $\mathcal{Y}$>, a randomly initialized encoder-decoder model $\mathcal{M}=(f,g)$; return a contrastive generation model.}
\label{alg:training}
\begin{algorithmic}[1]
\Procedure{WarmUp}{$\mathcal{M}$}
    \State Update the parameters of randomly initialized $\mathcal{M}$ with $\nabla_\theta\mathcal{L}_{nll}$ until convergence
\EndProcedure
\end{algorithmic}

\begin{algorithmic}[1]
\Procedure{BeamSearch}{$g$, $\mathbf{H}_X$, $b$}
    \algorithmiccomment{beam search algorithm}
    \State \textbf{return} Text, likelihood, logits of the $b$ hypotheses
\EndProcedure
\end{algorithmic}
\begin{algorithmic}[1]
\Procedure{Train}{$\mathcal{M}$, <$\mathcal{X}$, $\mathcal{Y}$>}
\State $\theta \leftarrow$ Parameters of $\mathcal{M}$, $b \leftarrow $ beam size
% $div \leftarrow $ diversity penalty
\State \textsc{WarmUp}($\mathcal{M}$)
\While{not convergence}
\State $X^{1:k},Y^{1:k} \leftarrow$ A minibatch of $k$ datapoints from  <$\mathcal{X}$, $\mathcal{Y}$>
\State$\mathbf{H}_X^{1:k} \leftarrow {f}\,(X^{1:k}),\, \mathbf{H}_Y^{1:k} \leftarrow {g}\,(\mathbf{H}_X^{1:k}, Y^{1:k})$ 
\algorithmiccomment{outputs from the encoder and decoder }
\State ${Y'}^{1:k, 1:b}, \mathbf{P}_{{Y}'}^{1:k, 1:b}, \mathbf{H}_{{Y'}}^{1:k, 1:b}$ = \textsc{BeamSearch$(g, \mathbf{H}_X^{1:k}, b)$}
\State ${Y'}^{1:k, 1:(b+k)} \leftarrow$  Append $b$ self-generated samples to $Y^{1:k}$ 
\For{$i \in 1,2,\ldots,k$}
\For{$j \in 1,2,\ldots,,b+k$}
\State Do oracle measurement $o({Y'}^{i,j}, Y^{i})$ for each element ${Y'}^{i,j}$  in ${Y'}^{1:k, 1:(b+k)}$
\EndFor
\EndFor
$\mathcal{L}_{ctr} \leftarrow$ Get pair-wise contrastive loss 
\State update parameters using $\nabla_\theta(\mathcal{L}_{nll}+\mathcal{L}_{ctr})$
\EndWhile
\State \textbf{return} $\mathcal{M}$
\EndProcedure
\end{algorithmic}
\end{algorithm}

\begin{figure}
\centering
\begin{subfigure}{.46\textwidth}
    \begin{center}
        \vspace{-0.5em}
        \includegraphics[width=\textwidth]{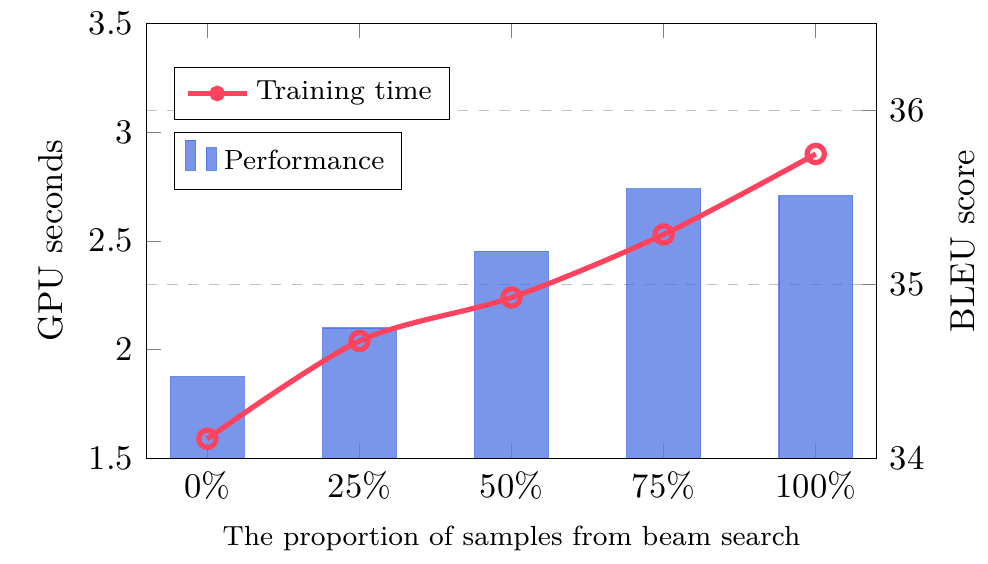}
        \caption{ Relationship of GPU seconds (left axis) and BLEU (right axis) with the proportion of self-generated samples on on the test set of IWSLT'14 De-En.  We set the maximum size of candidate samples to 32. }
        \label{fig:proportion}
    \end{center}
\end{subfigure}
\hspace{2mm}
\begin{subfigure}{.41\textwidth}
  \begin{center}
    % \vspace{1.9em}
    \includegraphics[width=\textwidth]{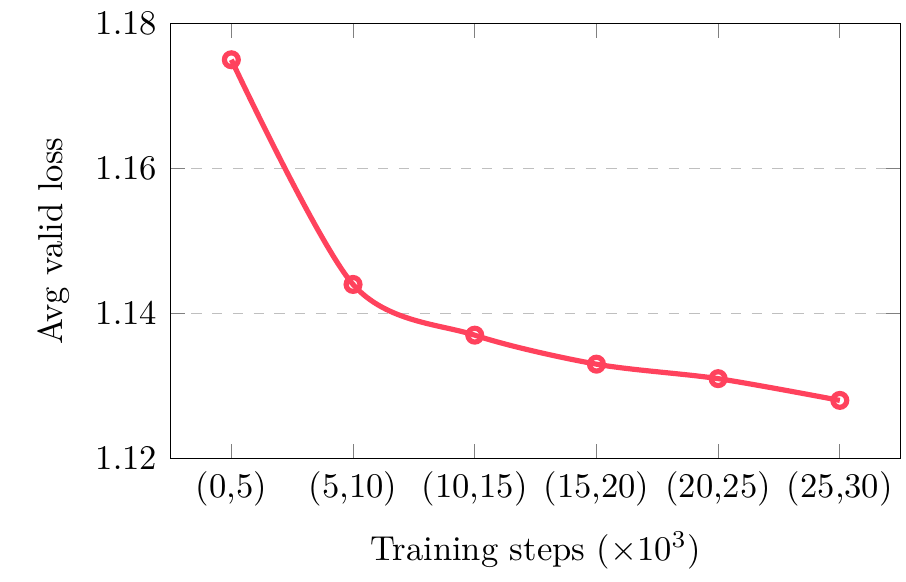}
    \caption{Relationship between the contrastive loss on IWSLT'14 De-En validation set and training steps. We calculate the validation loss every 1k steps and report the average results every 5k steps.}
    \label{fig:valid_loss}
  \end{center}
\end{subfigure}
\vspace{-0.5em}
\caption{ Analysis on using speed-accuracy trade-off tricks}
\vspace{-1em}
\end{figure}

\paragraph{Small Beam Size} The first trick is to reduce the proportion of self-generated samples. With the increase of beam size, the time consumed by beam search will increase significantly. Remaining the maximum number of contrastive examples for each input unchanged, we can adjust the ratio of self-generated samples and from-batch samples. For IWSLT'14 En-De benchmark with transformer small, the default maximum number of contrastive samples is set to 32 and the proportion of self-generated samples is 75\% (settings of other benchmarks are shown in Table~\ref{tab:cl-params}).  Relationship between the training time of each iteration and the proportion of samples returned by beam search on IWSLT'14 De-En benchmark can be seen in Figure~\ref{fig:proportion}\footnote{We run the two experiments on single NVIDIA Tesla A100 with maximum number of tokens per batch set to 4000 without gradient accumulation}. 
Totally using self-generated samples will double the training speed of the naive contrastive text generation method and will do not further boost the improvement in performance. 
Reducing the rate of self-generated samples to 50\% still resulting in 1.0 BLEU superior to the baseline while saving about 1.0 GPU seconds per iteration compared with totally using self-generated samples. 

\paragraph{Early Stop} Another way to save training time is early stop.  We can see in Figure~\ref{fig:valid_loss}, on IWSLT'14 De-En, the declining trend of the contrastive loss on validation set allowing us to perform early stop in training.  The contrastive loss on valdation set drop rapidly at the first 10k steps, and this decline will slow down in the following steps. In our experiments we train this model for about 40k steps, and early stop after 10k steps is also enough to improve the MLE baseline by 0.8 BLEU but saving 3/4 training time. 

\section{Experimental Setup}\label{sec:exp-setting}
In this section, we will introduce more details about
experiments and datasets. Our experiments on IWSLT'14 De-En and WMT‘14 En-De use fairseq~\cite{ott-etal-2019-fairseq} framework. Experiments on other datasets are run with transformers~\cite{wolf2019huggingface}.  Table~\ref{tab:data} gives an overview of the number of instances in train/validation/test set and its source. The test set of Totto~\cite{parikh2020totto} and CommonGen~\cite{lin2020commongen} is invisible. We get these results by submitting our generation results to the leaderboards\footnote{\url{https://inklab.usc.edu/CommonGen/leaderboard.html}}\footnote{\url{https://github.com/google-research-datasets/ToTTo}}.

\renewcommand\arraystretch{1.0}
\begin{table}[h]
\begin{center}
\caption{The statistics of datasets we use in experiments.}
\label{tab:data}
\tabcolsep0.03 in
\begin{tabular}{lcccl}
\toprule
{\bf Datasets} & \bf Train(\#) & \bf Validation(\#) & \bf Test(\#) & \bf Source(\#) \\
\midrule
WMT'16 Ro-En& 610k & 2k &  2k &  Romanian-to-English translation\\
IWSLT'14 De-En& 160k & 7k & 7k & Germanish-to-English translation\\
WMT'14 En-De & \textbf{4.5M} & 3k & 3k & English-to-Germanish translation\\
XSum~\cite{narayan2018don} & 204k & 11k & 11k &  One-sentence summary of BBC news articles\\
Multi-News~\cite{fabbri2019multi}&45k & 5.6k &  5.6k & Long summary of multiple news articles\\
Java~\cite{lu2021codexglue} & 165k & 5k & 11k & Code comment for java\\
Python~\cite{fabbri2019multi}& 252k & 14k & 15k & Code comment for Python\\
WikiBio~\cite{lebret2016neural} & 582k & 73k & 73k & Description of a table from Wiki\\
Totto~\cite{parikh2020totto} & 121k & 7.7k & 7.7k & Description of a table from Wiki\\
CommonGen~\cite{lin2020commongen} & 67k & 4k &1.5k& A sentence containing all required concepts\\

\bottomrule
\end{tabular}
\end{center}
\end{table}

\subsection{Hyperparameters Brought by Contrastive Learning}
Compared with MLE based models, there are three additional hyperparameters introduced by \textsc{CoNT}. The first one is the maximum number of contrastive samples $m$ for an input sequence,  the second one is  margin strength $\gamma$ (defined in Section~\ref{ssec:npairsloss}) and the third one is  the balance factor $\alpha$. $\gamma$ is set to 0.01 for all datasets. We tune $\alpha$ on the validation set from $[0.2,0.3,0.5,0.7]$. We set the $m$ to 32 on IWSLT'14 and set $m$ to 16 on   other benchmarks considering of efficiency. Actually, increasing the number of contrastive samples will continuously boost the performance. 
On all benchmarks, the beam size for diverse beam search~\cite{vijayakumar2016diverse} used in training is $0.75*m$ and the number of groups of diverse beam search is the same with beam size. Details can be found in Table~\ref{tab:cl-params}. Before adding contrastive learning, we pretrain the generation model with only negative log likelihood loss until the validation loss no longer decreases. All experiments are done on 4 NVIDIA Tesla A100 GPUs.

\renewcommand\arraystretch{1.1}
\begin{table}[t]
\begin{center}
\caption{ Hyperparameters brought by contrastive learning at training and inference stage. 'scratch` means the transformer model without pre-training. '$\alpha$` is the balance factor between likelihood and sequence similarity and $m$ means the maximum number of contrastive samples during training. }
\label{tab:cl-params}
\tabcolsep0.04 in
\begin{tabular}{lccccccc}
\toprule
\multirow{2}{*}{\vspace{-2mm}\bf Datasets} & \multirow{2}{*}{\vspace{-2mm}\bf model}& & \multicolumn{2}{c}{\bf Inference} & & \multicolumn{2}{c}{\bf Training} \\
\cmidrule{4-5} \cmidrule{7-8} 
& & & $\alpha$  & beam size & &  $m$ & beam size  \\ 
\midrule
\multicolumn{8}{c}{\textit{Small-scale model}} \\
\midrule
WMT'16 Ro-En & scratch & & 0.5 & 12 & & 16 & 12\\
WMT'16 Ro-En & T5 & & 0.5 & 12 & & 16 & 12\\
IWSLT'14 De-En & scratch & & 0.5 & 12 & & 32 & 24 \\
XSum~\cite{narayan2018don} & T5 & & 0.5 & 8 & & 16 & 12 \\
Multi-News~\cite{fabbri2019multi} & T5 & & 0.5 & 8 & & 16 & 12 \\
WikiBio~\cite{lebret2016neural} & T5 & & 0.3 & 8 & & 16 & 12 \\
\midrule
\multicolumn{8}{c}{\textit{Base-scale model}} \\
\midrule
WMT'14 En-De & scratch & & 0.2 & 8 & & 16 & 12 \\
Java~\cite{lu2021codexglue} & CodeT5 & & 0.2 & 8 & & 16 & 12 \\
Python~\cite{lu2021codexglue} & CodeT5 & & 0.2 & 8 & & 16 & 12\\
Totto~\cite{parikh2020totto} & T5 & & 0.3 & 8 & & 16 & 12 \\
CommonGen~\cite{lin2020commongen} & T5 & & 0.2 & 4 & & 16 & 12 \\
\midrule
\multicolumn{8}{c}{\textit{Large-scale model}} \\
\midrule
XSum~\cite{narayan2018don} & Pegasus & & 0.3 & 12 & & 16 & 12 \\
Multi-News~\cite{fabbri2019multi} & Pegasus & & 0.5 & 4 & & 16 & 12\\
\bottomrule
\end{tabular}
\end{center}
\end{table}

\subsection{Machine Translation} 
For WMT'16 Ro-En, We use the Adafactor optimizer following~\cite{raffel2019exploring} to finetune transformer-small and T5-small with learning rate = $1\times10^{-3}$.  The pre-trained checkpoints  of  T5 are provided by transformers\footnote{\url{https://github.com/huggingface/transformers}} 
We limit the maximum input/output length to 128.   Validation step is performed every 2000 training steps. We train our model for 2 epochs and get the best model at step 8000. It takes about 2 hours on 4 NVIDIA Tesla A100 GPUs with a batch size of 32.
The dimension of hidden state of small-scale model is 512. So the dimension of representations from the encoder and decoder $\mathbf{z}$ is as the same. At inference stage we set the length penalty to 1.0.

Previous work mainly report their results on IWSLT'14 De-En and WMT'14 En-De based on fairseq~\cite{ott-etal-2019-fairseq} library\footnote{\url{https://github.com/facebookresearch/fairseq}}. We also implement \textsc{CoNT} on the two datasets with fairseq. 
For IWSLT'14 De-En, we use the small setup of the Transformer model. The model has 6 layers where model dimension for each layer is 512 and  feed-forward dimension is set to 1024. 
The batch size is up to 4000 tokens and we update our model every 4 backwards.
On WMT'14 En-De,  we use the base setup of the Transformer model where the dimension of feed-forward layer is set to 2048. The embedding for  decoder input and output is shared. The batch size is also set to 4000 tokens but we update every 20 backwards to simulate large batch size which is very crucial to WMT'14 benchmark. In addition, we find that  open the dropout module in decoder during inference will help the performance for \textsc{CoNT} on WMT'14 En-De.

We train our model for 20 epochs on IWSLT'14 De-En and 10 epochs on WMT'14 En-De. We use 4 GPUs for the model training. The average running time for IWSLT is about 8 hours and for WMT is around 32 hours.
We use FP16 to accelerate our training.
Other settings are the same with the default settings recommended by the instruction of fairseq official site to re-produce the neural machine translation
results~\footnote{\url{https://github.com/facebookresearch/fairseq/blob/main/examples/translation/README.md}}.
For both IWSLT'14 De-En and WMT'14 En-De, 
we use Adam optimizer with learning rate $5\times10^{-4}$  with 
the inverse sqrt learning rate scheduler to optimize the models.

\subsection{Summarization}
 We use the Adafactor optimizer with learning rate = $1\times10^{-3}$ to finetune T5-small model and Pegasus-large~\cite{zhang2020PEGASUS} model. For XSum~\cite{narayan2018don}, we limit the input length to 512 and output length to 64. The input length is extended to 1024 and the output length is extended to 300 for multi-document summarization benchmark multi-news. 
 The length penalty for XSum is set to 0.8 while for multi-news, which has longer target sequence, the length penalty is set to 2.0
 We use a batch size of 32 for small-scale model and 4 for large-scale model. We train our model until the validation loss do not decrease. The total training hours using 4 GPUs is about 6 hours for small model  and 12 hours for large model.
 
\subsection{Code Comment Generation}
We use the state-of-the-art code comment generation model CodeT5~\cite{wang2021codet5} as our base model. We download their pre-trained checkpoint from transformers.
we truncate the input length to 512 and output length to 64.
The two benchmark is sensitive to batch size and learning rate. Therefore, we use a smaller learning rate $1\times10^{-4}$ with Adafactor optimizer. The batch size is set to 8 and other settings are the same with the origin paper of CodeT5\footnote{\url{https://github.com/salesforce/CodeT5}}. We train \textsc{CoNT} on the two benchmark for about 4  hours on 4 GPUs. The length penalty is set to 0.6 during decoding.

\subsection{Data-to-text Generation}
The input of data-to-text generation tasks is structured data (e.g., table, graph). To input the structured data into a  sequence-to-sequence model, we should first the linear the input. We linear the input for WikiBio following \citet{liu2018table}\footnote{\url{https://github.com/tyliupku/wiki2bio}}. For totto, we use preprocess the dataset following the instruction of the official site\footnote{\url{https://github.com/google-research/language/tree/master/language/totto}}.  We use the T5-small model as base model for WikiBio and T5-base for Totto. We limit the input length to 512 and output length to 128. 
The length penalty for WikiBio and Totto is set to 2.0.
We train our model for about 24 hours on 4 GPUs with a batch size of 32. 

\subsection{Commonsense Generation}
For commonsense generation task, we use the popular benchmark CommonGen~\citet{lin2020commongen}. The input of CommonGen is a set of concepts and the output is a fluency sentence mentioning all concepts in the source side. We concatenate these concepts with ',` as separator.  We use the base setup of the T5 model for CommonGen~\cite{lin2020commongen}. The maximum input length for source and target is limited to 64.  Other settings are the same with the settings of Totto. We train our model for 1 epoch upon the checkpoint   pre-trained with negative log likelihood loss. Since the scale of the dataset is samll,  it only takes about 0.5 hours training to convergence on 4 GPUs.

\section{Case Study}
We show some randomly selected examples from  IWSLT'14 Germanish-to-English translation task and XSum which aims to summarize a news article in table~\ref{tab:nmt-examples},\ref{tab:summ-examples},\ref{tab:summ-examples2}.
\renewcommand\arraystretch{1.25}
\begin{table*}[h!]
\caption{ Generation results of IWSLT'14 Germanish-to-English translation task (base model: Transformer-small).}
\label{tab:nmt-examples}
    \centering
    \small
    \begin{tabular}{@{}c  p{0.7\textwidth}}
    \toprule
    \multicolumn{1}{l}{\bf Germanish:} & dann kann ich das ganze übertragen. \\
    \multicolumn{1}{l}{\bf Ground Truth:} & and then i can transfer the whole thing. \\
    \multicolumn{1}{l}{\bf \textsc{CoNT}:} & and then i can translate the whole thing. \\
    \multicolumn{1}{l}{\bf MLE:} & then i can transmit this whole thing.
 \\\cmidrule{1-2}
  \multicolumn{1}{l}{\bf Germanish:} &sie machen sich immer sorgen, dass sie regalfläche verlieren.\\
    \multicolumn{1}{l}{\bf Ground Truth:} & they‘re always worried they’re going to lose shelf space. \\
    \multicolumn{1}{l}{\bf \textsc{CoNT}:} & they‘re always worried that they’re losing the shelf.\\
    \multicolumn{1}{l}{\bf MLE:} & they always worry that they lose real galleries.
 \\\cmidrule{1-2}
    \multicolumn{1}{l}{\bf Germanish:} & gewissermassen überflügelt uns die technik also. \\
    \multicolumn{1}{l}{\bf Ground Truth:} & so in a sense, it's getting ahead of us. \\
    \multicolumn{1}{l}{\bf \textsc{CoNT}:} & so, in a sense, technology overwhelms us. \\
    \multicolumn{1}{l}{\bf MLE:} & so, to some extent, technology overrivers us.
%  \\\cmidrule{1-2}
%   \multicolumn{1}{l}{\bf Germanish:} & es ist' ne form von revierverhalten, ja? \\
%     \multicolumn{1}{l}{\bf Ground Truth:} & it's a type of territorial behavior, right? \\
%     \multicolumn{1}{l}{\bf \textsc{CoNT}:} & it's the ne form of refourth behavior, right? \\
%     \multicolumn{1}{l}{\bf MLE:} & it's real-life form, right?
 \\\cmidrule{1-2}
   \multicolumn{1}{l}{\bf Germanish:} & erzählen sie mir über die "warum" phase--was bringt sie uns?\\
    \multicolumn{1}{l}{\bf Ground Truth:} & tell me about the  "why" phase -- what does that do for us?\\
    \multicolumn{1}{l}{\bf \textsc{CoNT}:} & tell me about the  "why" phase -- what does it bring us? \\
    \multicolumn{1}{l}{\bf MLE:} &tell me about why -- what does it bring us?
 \\\cmidrule{1-2}
   \multicolumn{1}{l}{\bf Germanish:} & und wir können auf diese sehr einfache weise navigieren. \\
    \multicolumn{1}{l}{\bf Ground Truth:} & and we can just navigate in this very simple way. \\
    \multicolumn{1}{l}{\bf \textsc{CoNT}:} & and we can navigate in this very simple way. \\
    \multicolumn{1}{l}{\bf MLE:} & and we can navigate this very simple way.
\\\cmidrule{1-2}
   \multicolumn{1}{l}{\bf Germanish:} & und wenn ich jemandem sage, gib mir mal salz oder pfeffer, dann wird er beim rechten erstmal überlegen, wo ist was drin.\\
    \multicolumn{1}{l}{\bf Ground Truth:} & and if i say to someone, pass me the salt or the pepper, they'd have to first think about what's in what with the right one. \\
    \multicolumn{1}{l}{\bf \textsc{CoNT}:} & and when i say to someone, i'll give me salt or pepper, they'll put it on the right for the first time, where is it in. \\
    \multicolumn{1}{l}{\bf MLE:} & and when i tell someone, give me salt or pepper, they'll be thinking about where is there?
 \\\cmidrule{1-2}
    \multicolumn{1}{l}{\bf Germanish:} & und problem , das ist nicht nur ein technisches problem, es kann auch ein gesellschaftliches problem sein, es kann auch einfach ein zugangsproblem sein , was wie dinge vereinfachen, also eine beliebige problemstellung , eine frage aufzuwerfen, und wie kann man das anders oder wie kann man das besser machen. \\
    \multicolumn{1}{l}{\bf Ground Truth:} & and a problem, so not only a technical problem, it can also be a social problem, it can also just be an access problem that simplifying things, so any way of looking at a problem, of posing a question, asking how you could do something differently or better. \\
    \multicolumn{1}{l}{\bf \textsc{CoNT}:} & and the problem is, it‘s not just a technical problem, it can also be a social problem, it can be a problem as well as a problem of accessing how things simplify, which is an arbitrary problem of asking a question, and how do you do it differently, or how do you do it better. \\
    \multicolumn{1}{l}{\bf MLE:} & and the problem is, it‘s not just a technical problem, it can also be a social problem, it can be a problem as well as a problem of accessing how things simplify, which is an arbitrary problem of asking a question, and how do you do it differently, or better. \\
\bottomrule
\end{tabular}
\end{table*}
\renewcommand\arraystretch{1.2}
\begin{table*}[h!]
\caption{ Generation results of XSum (base model: T5-small).}
\label{tab:summ-examples}

    \centering
    \small
    \begin{tabular}{@{}c  p{0.7\textwidth}}
    \toprule
      \multicolumn{1}{l}{\bf Document:} &  The 23-year-old from Guernsey appointed Veronelli in December 2013, but he is no longer able to commit to spending up to 40 weeks a year on the road. Veronelli, 36, moved from Florida back to Buenos Aires earlier this year to be with his young family. Watson, the world number 55, won her second WTA tour title at the Hobart International in January. Veronelli, a former world number 150, had notable success in guiding Watson back inside the world's top 50 for a time, after she had slipped down the rankings following a bout of glandular fever in 2013. \\
    \multicolumn{1}{l}{\bf Ground Truth:} &  British number two Heather Watson has parted company with her Argentine coach Diego Veronelli.\\
    \multicolumn{1}{l}{\bf \textsc{CoNT}:} & British number two Tom Watson has withdrawn from the WTA Tour due to illness.\\
    \multicolumn{1}{l}{\bf MLE:} & World number one Laura Watson has been reunited with his wife, Veronelli, after a long illness.
     \\\cmidrule{1-2}
    \multicolumn{1}{l}{\bf Document:} & The 26-year-old was released by York City after failing to score in 14 appearances last season. However, he netted 26 times in 69 appearances in a two-season spell at Barnet between 2012 and 2014. Hyde is the Boro's fourth signing of the summer, following left-back Andrew Fox and forwards Matt Godden and Rowan Liburd. "I know this league inside out now and any team can go on a run, but it's who does it for the longest period that counts. "It's about winning games and fingers crossed I can help Stevenage do that this season," he told the club website. Details of his contract with Stevenage have not been disclosed. Find all the latest football transfers on our dedicated page. \\
    \multicolumn{1}{l}{\bf Ground Truth:} & League Two side Stevenage have signed their third striker of the summer by bringing in free agent Jake Hyde. \\
    \multicolumn{1}{l}{\bf \textsc{CoNT}:} & League Two side Stevenage have signed striker Jordan Hyde on a two-year contract.\\
    \multicolumn{1}{l}{\bf MLE:} &  Stevenage have signed forward Ryan Hyde on a two-year deal. Accrington Stanley loan deal.
     \\\cmidrule{1-2}

     \multicolumn{1}{l}{\bf Document:} &Both sides had chances before the Pars' Ryan Wallace drilled a low shot into the bottom corner with 15 minutes left. The lead last just two minutes as Ross Davidson got the last touch on a free-kick into the area. The home side dominated the closing stages but could could not deny Rovers, who remain in fifth place. Rovers remain level with Airdrieonians, who drew at home with Forfar Athletic.  \\
    \multicolumn{1}{l}{\bf Ground Truth:} &  Scottish League One leaders Dunfermline Athletic were held at home by Albion Rovers but still moved 11 points clear at the top of the table. \\
    \multicolumn{1}{l}{\bf \textsc{CoNT}:} & Tranmere Rovers slipped to the bottom of Scottish League One as they were held to a draw by play-off chasing Pars. \\
    \multicolumn{1}{l}{\bf MLE:} & Dundee Rovers dominated the Scottish Championships with a 2-0 win over Forfar Athletic.
      \\\cmidrule{1-2}
     \multicolumn{1}{l}{\bf Document:} &Reports in France suggest Toulon coach Diego Dominguez's job is under threat, and Lancaster, 46, is viewed as a potential successor. He left his role with England after their exit from the 2015 World Cup. Since his departure, Lancaster has been an adviser to the Football Association, the NFL and British Cycling. He is interested in the Toulon job, but is understood to still be keen on a role in the southern hemisphere, with posts in Australia at the Queensland Reds and the Western Force both available.\\
    \multicolumn{1}{l}{\bf Ground Truth:} & Former England boss Stuart Lancaster is meeting Toulon president Mourad Boudjellal this week as he seeks a return to full-time coaching. \\
    \multicolumn{1}{l}{\bf \textsc{CoNT}:} & Former England coach Stuart Lancaster is interested in a role in the southern hemisphere, according to the Football Association. \\
    \multicolumn{1}{l}{\bf MLE:} & England's Chris Lancaster has been appointed as the new head coach of the Toulon Football Association.
\\\bottomrule
\end{tabular}
\end{table*}

\renewcommand\arraystretch{1.1}
\begin{table*}[h!]
\caption{ Generation results of XSum (base model: T5-small).}
\label{tab:summ-examples2}
    \centering
    \small
    \begin{tabular}{@{}c  p{0.7\textwidth}}
    \toprule
     \multicolumn{1}{l}{\bf Document:} &
    The men were believed to have been working on the northbound carriageway of the A361 at Gornhay Cross, Tiverton, when they were hit. One was flown to Plymouth's Derriford Hospital, while a second Devon Air Ambulance flew the other man to the Royal Devon and Exeter Hospital. The van driver has been arrested on suspicion of dangerous driving. The condition of the two injured men, who are both in their 40s, is not known. Devon and Cornwall Police said a third man was hurt in the incident, but he is described as the "walking wounded". The Barnstaple-bound carriageway is expected to remain closed for several hours and diversions have been put in place.\\
     \multicolumn{1}{l}{\bf Ground Truth:} & ref:  Two road workers have been seriously hurt in an accident involving a van in mid Devon.\\
    \multicolumn{1}{l}{\bf \textsc{CoNT}:} &Two men have been seriously injured in a collision involving a van in Devon.\\
    \multicolumn{1}{l}{\bf MLE:} & A man has been arrested on suspicion of dangerous driving after two men were hit by a van on the M4 in Devon.
     \\\cmidrule{1-2}

    \multicolumn{1}{l}{\bf Document:} & Officers were called at about 02:30 BST following reports that a man in his 30s had been attacked in Greenbrow Road, Wythenshawe. He was taken to hospital but he died from his injuries. Supt Steve Howard, from Greater Manchester Police, said: "We are working hard to piece together what happened to the man." \\
    \multicolumn{1}{l}{\bf Ground Truth:} & A murder investigation has been launched after a man was found stabbed in Manchester, police have said.\\
    \multicolumn{1}{l}{\bf \textsc{CoNT}:} & A murder inquiry has been launched after a man was stabbed to death in Manchester. \\
    \multicolumn{1}{l}{\bf MLE:} & An 81-year-old man has died after being attacked in a street in Greater Manchester.  
     \\\cmidrule{1-2}
     \multicolumn{1}{l}{\bf Document:} & Damage to to overhead wires meant the line is blocked north of Morpeth. Virgin East Coast, Northern Rail, and Cross Country services were affected, with reports of large queues at Newcastle Central Station. Buses were organised to take passengers between Newcastle and Edinburgh, with people advised to avoid travelling if possible. Services resumed late on Friday.\\
    \multicolumn{1}{l}{\bf Ground Truth:} & Rail passengers travelling between Newcastle and Scotland faced severe disruption on Friday. \\
    \multicolumn{1}{l}{\bf \textsc{CoNT}:} &  Rail services between Newcastle and Edinburgh have been disrupted after a power cut led to delays. \\
    \multicolumn{1}{l}{\bf MLE:} & Trains across the UK have been cancelled due to a disruption to the main line in the Highlands.
    \\\cmidrule{1-2}
     \multicolumn{1}{l}{\bf Document:} & Rashan Charles, 20, was wrestled to the ground in Dalston, east London, on 22 July, and died about an hour later. On Friday, clashes broke out in Hackney as protesters blocked part of Kingsland Road and set mattresses alight. A spokesman for Mr Charles's family said they understood the anger but called for "dignified" protest. "Burning down homes will not give justice," he said. Mr Charles was pursued by officers and became ill after trying to swallow an object, the Met has said. He died soon after in hospital. The Independent Police Complaints Commission is investigating. Police warned that anyone using Mr Charles's death "as an excuse to commit crime" would be "dealt with robustly". Appealing for calm, family spokesman Stafford Scott said: "We understand your frustration, we understand your anger - don't feel that the family doesn't feel the anger and the frustration too. "But what the family knows is taking it to the streets doesn't give you justice. "Burning down your own homes, burning down your neighbourhood is not going to give you justice." Mr Scott, who runs race advocacy group Tottenham Rights... \\
    \multicolumn{1}{l}{\bf Ground Truth:} & The family of a black man who died after being apprehended by police has appealed for peace after violent protests in the wake of his death.\\
    \multicolumn{1}{l}{\bf \textsc{CoNT}:} & The family of a man who died after he was attacked by anti-racism protesters have appealed for calm.\\
    \multicolumn{1}{l}{\bf MLE:} & nll:  Thousands of black people have protested against the death of a man who was killed in a street attack.
\\\bottomrule
\end{tabular}
\end{table*}

\end{document}